\documentclass[11pt,twoside]{article}

\usepackage{fullpage}
\usepackage{amsfonts}

\usepackage{epsf}
\usepackage{fancyheadings}
\usepackage{graphics}
\usepackage{graphicx}

\usepackage{subcaption}
\usepackage{psfrag}

\usepackage{color}

\usepackage{amsthm}
\usepackage{amsmath}
\usepackage{amssymb}
\usepackage[linesnumbered, ruled, vlined]{algorithm2e}

\theoremstyle{plain}

\newtheorem{theorem}{Theorem}

\newtheorem{lemma}{Lemma}

\newlength{\widebarargwidth}
\newlength{\widebarargheight}
\newlength{\widebarargdepth}

\makeatletter
\long\def\@makecaption#1#2{
        \vskip 0.8ex
        \setbox\@tempboxa\hbox{\small {\bf #1:} #2}
        \parindent 1.5em  %% How can we use the global value of this???
        \dimen0=\hsize
        \advance\dimen0 by -3em
        \ifdim \wd\@tempboxa >\dimen0
                \hbox to \hsize{
                        \parindent 0em
                        \hfil 
                        \parbox{\dimen0}{\def\baselinestretch{0.96}\small
                                {\bf #1.} #2
                                } 
                        \hfil}
        \else \hbox to \hsize{\hfil \box\@tempboxa \hfil}
        \fi
        }
\makeatother

\long\def\comment#1{}

\newcommand{\vecnorm}[2]{\| #1\|_{#2}}

\newcommand{\defn}{:\,=}

\newcommand{\id}{\mathsf{id}}

\newcommand{\KT}{\mathsf{KT}}

\newcommand{\csst}{\mathbb{C}_{\mathsf{SST}}}
\newcommand{\cdiff}{\mathbb{C}_{\mathsf{DIFF}}}

\newcommand{\cbiso}{\mathbb{C}_{\mathsf{BISO}}}
\newcommand{\Cns}{\mathbb{C}_{\mathsf{NS}}}

\newcommand{\Mns}{M_{\mathsf{NS}}}

\newcommand{\bl}{\mathsf{bl}}
\newcommand{\Bl}{\mathsf{B}}
\newcommand{\Rw}{\mathsf{R}}

\newcommand{\tauhat}{\widehat{\tau}}
\newcommand{\Chatt}{\widehat{C}_t}

\newcommand{\rhat}{\widehat{\mathsf{r}}}
\newcommand{\rspace}{\mathsf{r}}

\newcommand{\elts}{\mathsf{s}}

\newcommand{\lambdahat}{\widehat{\lambda}}

\newcommand{\Cmodel}{\mathbb{C}}

\newcommand{\sgn}{\mathsf{sgn}}

\newcommand{\E}{\ensuremath{{\mathbb{E}}}}

\newcommand{\Mrisk}{\ensuremath{\mathcal{M}}}
\newcommand{\Erisk}{\ensuremath{\mathcal{E}}}
\newcommand{\Arisk}{\ensuremath{\mathcal{A}}}

\newcommand{\1}{\ensuremath{{\sf (i)}}}
\newcommand{\2}{\ensuremath{{\sf (ii)}}}
\newcommand{\3}{\ensuremath{{\sf (iii)}}}
\newcommand{\4}{\ensuremath{{\sf (iv)}}}

\DeclareMathOperator{\Var}{var}

\newcommand{\BER}{\ensuremath{\mbox{Ber}}}

\newcommand{\Ospace}{\ensuremath{\mathcal{O}}}

\newcommand{\Deltatilde}{\ensuremath{\widetilde{\Delta}}}

\newcommand{\Mhat}{\ensuremath{\widehat{M}}}
\newcommand{\Mtilde}{\ensuremath{\widetilde{M}}}
\newcommand{\bhat}{\ensuremath{\widehat{b}}}
\newcommand{\Mhatasp}{\ensuremath{\widehat{M}}_{\mathsf{ASP}}}
\newcommand{\ASP}{\ensuremath{\mathsf{ASP}}}

\newcommand{\BAP}{\ensuremath{\mathsf{BAP}}}
\newcommand{\Mhatbap}{\ensuremath{\widehat{M}}_{\mathsf{BAP}}}

\newcommand{\real}{\ensuremath{\mathbb{R}}}

\newcommand{\pihat}{\ensuremath{\widehat{\pi}}}
\newcommand{\pihatasp}{\ensuremath{\widehat{\pi}_{\mathsf{ASP}}}}

\newcommand{\EE}{\ensuremath{\mathbb{E}}}
\newcommand{\numitems}{\ensuremath{n}}

\begin{document}

\begin{center}

{\bf{\LARGE{Worst-case vs Average-case Design for \\
Estimation from Fixed Pairwise Comparisons}}}

\vspace*{.2in}

{\large{
\begin{tabular}{ccc}
Ashwin Pananjady$^\dagger$ & Cheng Mao$^\star$ & Vidya Muthukumar$^\dagger$ \\
\end{tabular}
}}
{\large{
\begin{tabular}{cc}
Martin J. Wainwright$^{\dagger, \ddagger}$ & Thomas A. Courtade$^\dagger$ \\
\end{tabular}
}}
\vspace*{.2in}

\begin{tabular}{c}
Department of Electrical Engineering and Computer Sciences, UC Berkeley$^\dagger$ \\
Department of Statistics, UC Berkeley$^\ddagger$ \\
Department of Mathematics, MIT $^\star$ \\
\end{tabular}

\vspace*{.2in}

\today

\end{center}
\vspace*{.2in}
%%%%%%%%%%%%%%%%%%%%%%%%%%%%%%%%

\begin{abstract}
  Pairwise comparison data arises in many domains, including
  tournament rankings, web search, and preference elicitation.  Given
  noisy comparisons of a fixed subset of pairs of items, we study the
  problem of estimating the underlying comparison probabilities under
  the assumption of strong stochastic transitivity (SST).  We also
  consider the noisy sorting subclass of the SST model. We show that
  when the assignment of items to the topology is arbitrary, these
  permutation-based models, unlike their parametric counterparts, do
  not admit consistent estimation for most comparison topologies used
  in practice. We then demonstrate that consistent estimation is
  possible when the assignment of items to the topology is randomized,
  thus establishing a dichotomy between worst-case and average-case
  designs. We propose two estimators in the average-case setting and
  analyze their risk, showing that it depends on the comparison
  topology only through the degree sequence of the topology. The rates
  achieved by these estimators are shown to be optimal for a large
  class of graphs. Our results are corroborated by simulations on
  multiple comparison topologies.
\end{abstract}

\section{Introduction} \label{sec:intro}

The problems of ranking and estimation from ordinal data arise in a
variety of disciplines, including web search and information
retrieval~\cite{dwork2001rank}, crowdsourcing~\cite{chen2013pairwise},
tournament play~\cite{herbrich2006trueskill}, social choice
theory~\cite{caplin1991aggregation} and recommender
systems~\cite{baltrunas2010group}. The ubiquity of such datasets stems
from the relative ease with which ordinal data can be obtained, and
from the empirical observation that using pairwise comparisons as a
means of data elicitation can lower the noise level in the
observations~\cite{barnett2003modern, stewart2005absolute}.

Given that the number of items $\numitems$ to be compared can be very
large, it is often difficult or impossible to obtain comparisons
between all ${\numitems \choose 2}$ pairs of items. A subset of pairs
to compare, which defines the \emph{comparison topology}, must
therefore be chosen. For example, such topologies arise from
tournament formats in sports, experimental designs in psychology set
up to aid interpretability, or properties of the elicitation process.
For instance, in rating movies, pairwise comparisons between items of
the same genre are typically more abundant than comparisons between
items of dissimilar genres.  For these reasons, studying the
performance of ranking algorithms based on fixed comparison topologies
is of interest. Fixed comparison topologies are also important in rank
breaking~\cite{hajek2014minimax, khetan2016data}, and more generally
in matrix completion based on structured
observations~\cite{kiraly2015algebraic, pimentel2016characterization}.

An important problem in ranking is the design of accurate models for
capturing uncertainty in pairwise comparisons.  Given a collection of
$n$ items, the results of pairwise comparisons are completely
characterized by the $n$-dimensional matrix of comparison
probabilities,\footnote{A comparison probability refers to the
  probability that item $i$ beats item $j$ in a comparison between
  them.} and various models have been proposed for such matrices.  The
most classical models, among them the
Bradley-Terry-Luce~\cite{bradley52rank,luce59individual} and Thurstone
models~\cite{thurstone27law}, assign a quality vector to the set of
items, and assign pairwise probabilities by applying a cumulative
distribution function to the difference of qualities associated to the
pair.  There is now a relatively large body of work on methods for
ranking in such parametric models (e.g., see the
papers~\cite{negahban2016rank, hajek2014minimax, chen2015spectral,
  shah16estimation} as well as references therein).  In contrast, less
attention has been paid to a richer class of models proposed decades
ago in the sociology literature~\cite{fishburn1973binary,
  mclaughlin1965stochastic}, which impose a milder set of constraints
on pairwise comparison matrix. Rather than positing a quality vector,
these models impose constraints that are typically given in terms of a
latent permutation that rearranges the matrix into a specified form,
and hence can be referred to as \emph{permutation-based} models. Two
such models that have been recently analyzed are those of strong
stochastic transitivity~\cite{shah17stochastically}, as well as the
special case of noisy sorting~\cite{braverman08noisy}. The strong
stochastic transitivity (SST) model, in particular, has been shown to
offer significant robustness guarantees and provide a good fit to many
existing datasets~\cite{ballinger1997decisions}, and this flexibility
has driven recent interest in understanding its properties. Also,
perhaps surprisingly, past work has shown that this additional
flexibility comes at only a small price when one has access to all
possible pairwise comparisons, or more generally, to comparisons
chosen at random~\cite{shah17stochastically}; in particular, the rates
of estimation in these SST models differ from those in parametric
models by only logarithmic factors in the number of items.  On a
related note, permutation-based models have also recently been shown
to be useful in other settings like
crowd-labeling~\cite{shah2016permutation}, statistical
seriation~\cite{flammarion16optimal} and linear
regression~\cite{pananjady2016linear}.

Given pairwise comparison data from one of these models, the problem
of estimating the comparison probabilities has applications in
inferring customer preferences in recommender systems, advertisement
placement, and sports, and is the main focus of this paper.

%%%%%%%%%%%%%%%%%%%%%%%%%%%%%%%%%%%%%%%%%%%%%%%%%%%%%%%%%%%%%%%%%%%%%%%

\paragraph{Our Contributions:}

Our goal is to estimate the matrix of comparison probabilities for
fixed comparison topologies, studying both the noisy sorting and SST
classes of matrices. Focusing first on the worst-case setting in which
the assignment of items to the topology may be arbitrary, we show in
Theorem~\ref{thm:minimax} that consistent estimation is impossible for
many natural comparison topologies. This result stands in sharp
contrast to parametric models, and may be interpreted as a ``no free
lunch'' theorem: although it is possible to estimate SST models at
rates comparable to parametric models when given a full set of
observations~\cite{shah17stochastically}, the setting of fixed
comparison topologies is problematic for the SST class.  This can be
viewed as a price to be paid for the additional robustness afforded by
the SST model.

Seeing as such a worst-case design may be too strong for
permutation-based models, we turn to an average-case setting in which
the items are assigned to a fixed graph topology in a randomized
fashion. Under such an observation model, we propose and analyze two
efficient estimators: Theorems~\ref{thm:ubeff} and~\ref{thm:sstupper}
show that consistent estimation is possible under commonly used
comparison topologies. Moreover, the error rates of these estimators
depend only on the degree sequence of the comparison topology, and are
shown to be unimprovable for a large class of graphs, in
Theorem~\ref{thm:lb1}.
 
Our results therefore establish a sharp distinction between worst-case
and average-case designs when using fixed comparison topologies in
permutation-based models. Such a phenomenon arises from the difference
between minimax risk and Bayes risk under a uniform prior on the
ranking, and may also be worth studying for other ranking models.

%%%%%%%%%%%%%%%%%%%%%%%%%%%%%%%%%%%%%%%%%%%%%%%%%%%%%%%%%%%%%%%%%%%%%%%%%%%%%
\paragraph{Related Work:}

The literature on ranking and estimation from pairwise comparisons is
vast, and we refer the reader to some
surveys~\cite{fligner1993probability, marden1996analyzing,
  cattelan2012models} and references therein for a more detailed
overview. Estimation from pairwise comparisons has been analyzed under
various metrics like top-$k$
ranking~\cite{chen2015spectral,shah2015simple,jang2016top,
  chen2017competitive} and comparison probability or parameter
estimation~\cite{hajek2014minimax,shah16estimation,shah17stochastically}.
There have been studies of these problems under
active~\cite{jamieson2011active,heckel2016active, maystre2015robust},
passive~\cite{negahban2016rank,rajkumar2016when}, and collaborative
settings~\cite{park2015preference, negahban2017learning}, and also for
fixed as well as random comparison
topologies~\cite{wauthier2013efficient,shah17stochastically}. Here we
focus on the subset of papers that are most relevant to the work
described here.

The problem of comparison probability estimation under a passively
chosen fixed topology has been analyzed for parametric models by Hajek
et al.~\cite{hajek2014minimax} and Shah et
al.~\cite{shah16estimation}. Both papers analyze the worst-case design
setting in which the assignment of items to the topology may be
arbitrary, and derive bounds on the minimax risk of parameter (or
equivalently, comparison probability) estimation. While their
characterizations are not sharp in general, the rates are shown to
depend on the spectrum of the Laplacian matrix of the topology. We
point out an interesting consequence of both results: in the
parametric model, provided that the comparison graph $G$ is connected,
the maximum likelihood solution, in the limit of infinite samples for
each graph edge, allows for exact recovery of the quality vector, and
hence matrix of comparison probabilities.  We will see that this
property no longer holds for the SST models considered in this paper:
there are comparison topologies and SST matrices for which it is
impossible to recover the full matrix even given an infinite amount of
data per graph edge.  It is also worth mentioning that the top-$k$
ranking problem has been analyzed for parametric models under fixed
design assumptions~\cite{jang2016top}, and here as well, asymptotic
consistency is observed for connected comparison topologies.

\vspace{-3mm}

\paragraph{Notation:}
Here we summarize some notation used throughout the remainder of this
paper.  We use $n$ to denote the number of items, and adopt the
shorthand $[n] \defn \{1, 2, \ldots, n\}$. We use $\BER(p)$ to denote
a Bernoulli random variable with success probability $p$. For two
sequences $\{a_n\}_{n=1}^\infty$ and $\{b_n\}_{n=1}^\infty$, we write
$a_n \lesssim b_n$ if there is a universal constant $C$ such that $a_n
\leq C b_n$ for all $n \geq 1$. The relation $a_n \gtrsim b_n$ is
defined analogously, and we write $a_n \asymp b_n$ if the relations
$a_n \lesssim b_n$ and $a_n \gtrsim b_n$ hold simultaneously. We use
$c, c_1, c_2$ to denote universal constants that may change from line
to line.

We use $\mathbf{e} \in \real^\numitems$ to denote the all-ones vector
in $\real^n$. Given a matrix $M \in \real^{n \times n}$, its $i$-th
row is denoted by $M_i$. For a graph $G$ with edge set $E$, let $M(G)$
denote the entries of the matrix $M$ restricted to the edge set of
$G$, and let $\|M\|_E^2 = \sum_{(i,j) \in E} M_{ij}^2$. For a matrix
$M \in \real^{n\times n}$ and a permutation $\pi:[n] \to [n]$, we use
the shorthand $\pi(M) = \Pi M \Pi^\top$, where $\Pi$ represents the
row permutation matrix corresponding to the permutation $\pi$. We let
$\id$ denote the identity permutation.  The Kendall tau
distance~\cite{kendall1948rank} between two permutations $\pi$ and
$\pi'$ is given by
\begin{align*} 
\KT(\pi,\pi') \defn \sum_{i,j \in [n]} \mathbf{1} \{\pi(i)< \pi(j),
\pi'(i)> \pi'(j) \}.
\end{align*}

Let $\mathcal{C}(G)$ represent the set of all connected,
vertex-induced subgraphs of a graph $G$, and let $V(S)$ and $E(S)$
represent the vertex and edge set of a subgraph $S$, respectively. We
let $\alpha(G)$ denote the size of the largest independent set of the
graph $G$, which is a largest subset of vertices that have no edges
among them. Define a biclique of a graph as two disjoint subsets of
its vertices $V_1$ and $V_2$ such that $(u,v) \in E(G)$ for all $u \in
V_1$ and $v \in V_2$. Define the biclique number $\beta(G)$ as the
maximum number of edges in any such biclique, given by $\max
\limits_{V_1, V_2 \text{ biclique}} |V_1||V_2|$. Let $d_v$ denote the
degree of vertex $v \in V$.

% Local Variables:
% TeX-master: "worstavg_design_arxiv"
% End:

\section{Background and Problem Formulation}
\label{sec:setup}

Consider a collection of $n \geq 2$ items that obey a total ordering
or ranking determined by a permutation $\pi^*:[n] \to [n]$.  More
precisely, item $i \in [n]$ is preferred to item $j \in [n]$ in the
underlying ranking if and only if $\pi^*(i) < \pi^*(j)$.  We are
interested in observations arising from stochastic pairwise
comparisons between items. We denote the matrix of underlying
comparison probabilities by $M^* \in [0,1]^{n \times n}$, with
$M^*_{ij} = \Pr\{i \succ j\}$ representing the probability that item
$i$ beats item $j$ in a comparison.

Each item $i$ is associated with a \emph{score}, given by the
probability that item $i$ beats another item chosen uniformly at
random. More precisely, the score $\tau^*_i$ of item $i$ is given by
\begin{align}
  \label{EqnDefnScore}
\tau^*_i & \defn [\tau(M^*)]_i \defn \frac{1}{n-1} \sum_{j \neq i}
M^*_{ij}.
\end{align}
Arranging the scores in descending order naturally yields a ranking of
items. In fact, for the models we define below, the ranking given by
the scores is consistent with the ranking given by $\pi^*$, i.e.,
$\tau_i \geq \tau_j$ if $\pi^*(i) < \pi^*(j)$. The converse also holds
if the scores are distinct.

%%%%%%%%%%%%%%%%%%%%%%%%%%%%%%%%%%%%%%%%%%%%%%%%%%%%%%%%%%%%%%%%%%%%%%%%%%%%

\subsection{Pairwise comparison models}
\label{sec:comparisonmodels}

We consider a permutation-based model for the comparison matrix $M^*$,
one defined by the property of \emph{strong stochastic
  transitivity}~\cite{fishburn1973binary, mclaughlin1965stochastic},
or the SST property for short.  In particular, a matrix $M^*$ of
pairwise comparison probabilities is said to obey the SST property if
for items $i, j$ and $k$ in the total ordering such that $\pi^*(i) <
\pi^*(j) < \pi^*(k)$, it holds\footnote{We set $M^*_{ii} = 1/2$
  by convention.}  that $\Pr(i \succ k) \geq \Pr(i \succ j)$.
Alternatively, recalling that $\pi(M)$ denotes the matrix obtained
from $M$ by permuting its rows and columns according to the
permutation $\pi$, the SST matrix class can be defined in terms of
permutations applied to the class $\cbiso$ of bivariate isotonic
matrices as
\begin{align}
  \label{EqnDefnSST}
\csst & \defn \bigcup_{\pi} \pi(\cbiso) = \bigcup_{\pi} \big\{\pi(M):
M \in \cbiso \big\}.
\end{align}
Here the class $\cbiso$ of bivariate isotonic matrices is given by
\begin{align*}
\{M \in [0,1]^{n \times n}: M + M^\top = \mathbf{e} \, \mathbf{e}^\top
\text{ and } M \text{ has non-decreasing rows and non-increasing
  columns} \},
\end{align*}
where $\mathbf{e} \in \real^n$ denotes a vector of all ones.

As shown by Shah et al.~\cite{shah17stochastically}, the SST class is
substantially larger than commonly used class of \emph{parametric}
models, in which each item $i$ is associated with a parameter $w_i \in
\mathbb{R}$, and the probability that item $i$ beats item $j$ is given
by $F(w_i - w_j)$, where $F: \mathbb{R} \mapsto [0,1]$ is a smooth
monotone function of its argument.

A special case of the SST model that we study in this paper is the
\emph{noisy sorting} model \cite{braverman08noisy}, in which the all
underlying probabilities are described with a single parameter
\mbox{$\lambda \in [0,1/2]$.}  The matrix $\Mns(\pi, \lambda) \in
     [0,1]^{n\times n}$ has entries
\begin{align*}
\big[\Mns(\pi, \lambda) \big]_{ij} = 1/2 + \lambda \cdot
\sgn\big(\pi(j) - \pi(i)\big),
\end{align*}
and the noisy sorting classes are given by
\begin{align}
\Cns(\lambda) \defn \bigcup_{\pi} \big\{\Mns(\pi, \lambda) \big\},
\quad \mbox{and} \quad \Cns \defn \bigcup_{\lambda \in [0,1/2]}
\Cns(\lambda). \label{eq:ns-def}
\end{align}
Here $\sgn(x)$ is the sign operator, with the convention that $\sgn(0)
= 0$. In words, the noisy sorting class models the case where the
probability $\Pr \{i \succ j\}$ depends only on the parameter
$\lambda$ and whether $\pi^*(i) < \pi^*(j)$. Although a noisy sorting
model is a very special case of an SST model, apart from the
degenerate case $\lambda^* = 1/2$, it cannot be represented by any
parametric model with a smooth function $F$, and so captures the
essential difficulty of learning in the SST class.

We now turn to describing the observation models that we consider in
this paper.

%%%%%%%%%%%%%%%%%%%%%%%%%%%%%%%%%%%%%%%%%%%%%%%%%%%%%%%%%%%%%%%%%%%%%%%%%%%%

\subsection{Partial observation models}

Our goal is to provide guarantees on estimating the underlying
comparison matrix $M^*$ when the comparison topology is fixed.
Suppose that we are given data for comparisons in the form of a graph
$G = (V, E)$, where the vertices represent the $n$ items and edges
represent the comparisons made between items. We assume that the
observations obey the probabilistic model
\begin{align}
  \label{eq:obs-fixed}
Y_{ij} =
\begin{cases}
\BER(M^*_{ij}) &\;\;\; \text{for } (i,j) \in E, \text{ independently}
\\ \star &\;\;\; \text{otherwise},
\end{cases}
\end{align}
where $\star$ indicates a missing observation. We
set the diagonal entries of $Y$ equal to $1/2$, and also specify that
$Y_{ji} = 1 - Y_{ij}$ for $j>i$, so that $Y+Y^\top = \mathbf{e} \,
\mathbf{e}^\top$.  We consider two different instantiations of the
edge set given the graph.

%%%%%%%%%%%%%%%%%%%%%%%%%%%%%%%%%%%%%%%%%%%%%%%%%%%%%%%%%%%%%%%%%%%%%%%%%%%%

\subsubsection{Worst-case setting}
\label{sec:ad}

In this setting, we assume that the assignment of items to vertices of
the comparison graph $G$ is arbitrary. In other words, once the graph
$G$ and its edges $E$ are fixed, we observe the entries of the matrix
according to the observation model \eqref{eq:obs-fixed}, and would
like to provide uniform guarantees in the metric $\|\Mhat - M^*\|_F^2$
over all matrices $M^*$ in our model class given this restricted set
of observations.

This setting is of the worst-case type, since the adversary is allowed
to choose the underlying matrix with knowledge of the edge set
$E$. Providing guarantees against such an adversary is known to be
possible for parametric models~\cite{hajek2014minimax,
  shah16estimation}. However, as we show in
Section~\ref{sec:adresults}, such a guarantee is impossible to obtain
even over the the noisy sorting subclass of the full SST
class. Consequently, the latter parts of our analysis apply to a
less rigid, average-case setting.

\subsubsection{Average-case setting}
\label{sec:unperm}

In this setting, we assume that the assignment of items to vertices of
the comparison graph $G$ is random. Equivalently, given a fixed comparison graph $G$ having adjacency matrix $A$, the subset of the entries that we observe can be modeled by the operator $\Ospace = \sigma(A)$ for a permutation $\sigma: [n] \to [n]$ chosen uniformly at random.
For a fixed
comparison matrix $M^*$, our observations themselves consist of a random subset of the entries of
the matrix $Y$ determined by the operator $\Ospace$: a location where $\mathcal{O}_{ij} = 1$ (respectively $\mathcal{O}_{ij} = 0$) indicates that entry $Y_{ij}$ is
observed (respectively is not observed). %The random observation
%operator $\mathcal{O}$ is determined by the graph, or more precisely,
%its adjacency matrix $A$: we set $\Ospace = \sigma(A)$, where $\sigma$
%is a permutation over $[n]$ chosen uniformly at random.  
Such a
setting is reasonable when the graph topology is constrained, but we
are still given the freedom to assign items to vertices of the
comparison graph, e.g. in psychology experiments. A natural extension
of such an observation model is the one of $k$ random designs,
consisting of multiple random observation operators $\{\Ospace_i =
\sigma_i(A)\}_{i=1}^k$, chosen with independent, random permutations
$\{\sigma_i\}_{i=1}^k$.

Our guarantees in the one sample setting with the observation operator
$\Ospace$ can be seen as a form of Bayes risk, where given a fixed
observation pattern $E$ (consisting of the entries of the comparison
matrix $Y$ determined by the adjacency matrix $A$ of the graph $G$,
with $A_{ij}$ representing the indicator that entry $Y_{ij}$ is
observed), we want to estimate a matrix $M^*$ under a uniform Bayesian
prior on the ranking $\pi^*$.  Studying this average-case setting is
well-motivated, since given fixed comparisons between a set of items,
there is no reason to assume a priori that the underlying ranking is
generated adversarially.

We are now ready to state the goal of the paper.  We address the
problems of recovering the ranking $\pi^*$ and estimating the matrix
$M^*$ in the Frobenius norm. More precisely, given the observation
matrix $Y=Y(E)$ (where the set $E$ is random in the average-case
observation model), we would like to output a matrix $\Mhat$ that is
function of $Y$, and for which good control on the Frobenius norm
error $\|\Mhat - M^*\|_F^2$ can be guaranteed.

% Local Variables:
% TeX-master: "worstavg_design_arxiv"
% End: 

\section{Main results} \label{sec:mainresults}

In this section, we state our main results and discuss some of their
consequences.  Proofs are deferred to Section~\ref{sec:proof}.

%%%%%%%%%%%%%%%%%%%%%%%%%%%%%%%%%%%%%%%%%%%%%%%%%%%%%%%%%%%%%%%%%%%%%%%%%%%

\subsection{Worst-case design: minimax bounds}
\label{sec:adresults}

In the worst-case setting of Section~\ref{sec:ad}, the performance of
an estimator is measured in terms of the normalized minimax error
\begin{align*}
\Mrisk(G, \Cmodel ) = \inf_{\Mhat = f(Y(G))} \sup_{M^* \in \Cmodel}
\EE \Big[ \frac{1}{n^2}\|\Mhat - M^* \|_F^2 \Big],
\end{align*}
where the expectation is taken over the randomness in the observations
$Y$ as well as any randomness in the estimator, and $\Cmodel \in
\{\csst, \Cns \}$ represents the model class.  Our first result shows
that for many comparison topologies, the minimax risk is prohibitively
large even for the noisy sorting model.

\begin{theorem}
  \label{thm:minimax}
For any graph $G$, the diameter of the set consistent with
observations on the edges of $G$ is lower bounded as
\begin{subequations}
\begin{align}
\sup_{\substack{M_1, M_2 \in \Cns \\ M_1(G) = M_2(G)}} \|M_1 - M_2
\|_F^2 \geq \alpha(G) (\alpha(G) - 1) \vee \beta(G^c). \label{eq:dia}
\end{align}
Consequently, the minimax risk of the noisy sorting model is lower
bounded as
\begin{align}
\Mrisk(G, \Cns) &\geq \frac{1}{4n^2} \left[ \alpha(G) (\alpha(G) - 1)
  \vee \beta(G^c) \right]. \label{eq:minimax}
\end{align}
\end{subequations}
\end{theorem}
Note that via the inclusion $\Cns \subset \csst$,
Theorem~\ref{thm:minimax} also implies the same lower
bound~\eqref{eq:minimax} on the risk $\Mrisk(G, \csst)$. In addition
to these bounds, the lower bounds for estimation in parametric models,
known from past work~\cite{shah16estimation}, carry over directly to
the SST model, since parametric models are subclasses of the SST class.

Theorem~\ref{thm:minimax} is approximation-theoretic in nature: more
precisely, inequality~\eqref{eq:dia} is a statement purely about the
size of the set of matrices consistent with observations on the
graph. Consequently, it does not capture the uncertainty due to noise,
and thus can be a loose characterization of the minimax risk for some
graphs, with the complete graph being one example.  The
bound~\eqref{eq:dia} on the diameter of the set of consistent
observations may be interpreted as the worst case error in the
infinite sample limit of observations on $G$. Hence,
Theorem~\ref{thm:minimax} stands in sharp contrast to analogous
results for parametric models~\cite{hajek2014minimax,
  shah16estimation}, in which it suffices for the graph to be
connected in order to obtain consistent estimation in the infinite
sample limit. For example, connected graphs with large independent
sets of order $n$ do not admit consistent estimation over the noisy
sorting and hence SST classes.

It is also worth mentioning that the connectivity properties of the
graph that govern minimax estimation in the larger SST model are quite
different from those appearing in parametric models. In particular,
the minimax rates for parametric models are closely related (via the
linear observation model) to the spectrum of the Laplacian matrix of
the graph $G$. In Theorem~\ref{thm:minimax}, however, we see other
functions of the graph appearing that are not directly related to
the Laplacian spectrum. In Section~\ref{sec:sim}, we evaluate these
functions for commonly used graph topologies, showing that for many
of them, the risk is lower bounded by a constant even for graphs
admitting consistent parametric estimation.

Seeing as the minimax error in the worst-case setting can be
prohibitively large, we now turn to evaluating practical estimators in
the random observation models of Section~\ref{sec:unperm}.

%%%%%%%%%%%%%%%%%%%%%%%%%%%%%%%%%%%%%%%%%%%%%%%%%%%%%%%%%%%%%%%%%%%%%%%%%%%%

\subsection{Average-case design: noisy sorting matrix estimation}
\label{sec:unpermresults}

In the average-case setting described in Section~\ref{sec:unperm}, we measure
the performance of an estimator using the risk
\begin{align*}
\sup_{M^* \in \Cmodel} \EE_{\Ospace, Y} \frac{1}{n^2} \| \Mhat - M^*
\|_F^2.
\end{align*}
It is important to note that the expectation is taken over both the
comparison noise, as well as the random observation pattern
$\Ospace$ (or equivalently, the underlying random permutation $\sigma$ assigning items to vertices). We propose the Average-Sort-Project estimator ($\ASP$ for short) for matrix estimation in
this metric, which is a natural generalization of the Borda count
estimator~\cite{chatterjee16estimation, shah16feeling}. It consists of
three steps, described below for the noisy sorting model:
\begin{enumerate}
\item[(1)] {\bf Averaging step:} Compute the average $\widehat{\tau}_i
  = \frac{\sum_{j \neq i} Y_{ij} \Ospace_{ij}}{\sum_{j \neq
      i} \Ospace_{ij}}$, corresponding to the fraction of comparisons
  won by item $i$.
\item[(2)] {\bf Sorting step:} Choose the permutation $\pihatasp$ such
  that the sequence $\{\tauhat_{\pihatasp^{-1}(i)}\}_{i=1}^n$ is
  decreasing in $i$, with ties broken arbitrarily.
\item[(3)] {\bf Projection step:} Find the maximum likelihood estimate
  $\lambdahat$ by treating $\pihatasp$ as the true permutation that
  sorts items in decreasing order. Output the matrix \mbox{$\Mhatasp
    \defn \Mns(\pihatasp, \lambdahat)$.}
\end{enumerate}

We now state an upper bound on the mean-squared Frobenius error
achievable using the $\ASP$ estimator.  It involves the degree
sequence $\{d_v\}_{v \in V}$ of a graph $G$ without isolated vertices,
meaning that $d_v \geq 1$ for all $v \in V$.
\begin{theorem}
  \label{thm:ubeff}
Let the observation process be given by $\mathcal{O}$. For any graph
$G = (V, E)$ without isolated vertices and any matrix \mbox{$M^* \in
  \Cns(\lambda^*)$}, we have
\begin{subequations}
  \begin{align}
 \label{eq:nsubeff} 
\EE_{\Ospace, Y} \left[ \frac{1}{n^2} \|\Mhatasp - M^* \|_F^2 \right]
& \lesssim \frac{1}{|E|} + \frac{n \log n}{|E|^2} +
\frac{\lambda^*}{n} \sum_{v \in V} \frac{1}{\sqrt{d_v}}, \qquad \text{
  and } \\
\label{eq:nsubkt}
\EE_{\Ospace, Y} \left[ \KT(\pi^*, \pihatasp) \right] &\lesssim
\frac{n}{\lambda^*} \sum_{v \in V} \frac{1}{\sqrt{d_v}}.
\end{align}
\end{subequations}
\end{theorem}
\noindent A few comments are in order.  First, while the results are
stated in expectation, a high probability bound can be proved for
permutation estimation---namely
\begin{align*}
\Pr_{\Ospace, Y} \Big\{\KT(\pi^*, \pihatasp) &\gtrsim \frac{n
  \sqrt{\log n}}{\lambda^*} \sum_{v \in V} \frac{1}{\sqrt{d_v}} \Big\}
\leq n^{-10}.
\end{align*}
Second, it can be verified that $\frac{1}{|E|} + \frac{n \log
  n}{|E|^2} \lesssim \frac{1}{n} \sum_{v \in V} \frac{1}{\sqrt{d_v}}$,
so that taking a supremum over the parameter $\lambda^* \in [0, 1/2]$
guarantees that the mean-squared Frobenius error is upper bounded as
$O\left( \frac{1}{n} \sum_{v \in V} \frac{1}{\sqrt{d_v}} \right)$,
uniformly over the entire noisy sorting class $\Cns$.  Third, it is
also interesting to note the dependence of the bounds on the noise
parameter $\lambda^*$ of the noisy sorting model. The ``high-noise''
regime $\lambda^* \approx 0$ is a good one for estimating the
underlying matrix, since the true matrix $M^*$ is largely unaffected
by errors in estimating the true permutation. However, as captured by
equation~\eqref{eq:nsubkt}, the permutation estimation problem is more
challenging in this regime.

The bound \eqref{eq:nsubeff} can be specialized to the complete graph
$K_{n}$ and the Erd\H{o}s-R\'enyi random graph with edge probability
$p$ to obtain the rates $1/\sqrt{n}$ and $1/\sqrt{np}$, respectively,
for estimation in the mean-squared Frobenius norm. These rates are strictly sub-optimal for these graphs,
since the minimax rates scale as $1/n$ and $1/(np)$, respectively;
both are achieved by the global MLE~\cite{shah17stochastically}. Such a
phenomenon is consistent with the gap observed between computationally
constrained and unconstrained estimators in similar and related
problems~\cite{shah17stochastically, flammarion16optimal,
  pananjady2017denoising}.

Interestingly, it turns out that the estimation
rate~\eqref{eq:nsubeff} is optimal in a certain sense, and we require
some additional notions to state this precisely.  Fix constants $C_1 = 10^{-2}$ and
$C_2 = 10^{2}$ and two sequences
$\{a_n\}_{n \geq 1}$ and $\{b_n\}_{n \geq 1}$ of (strictly) positive scalars.
For each $n \geq 1$, define the family of graphs
\begin{align*}
\mathcal{G}_n(a_n, b_n) & \defn \Big \{ G(V, E) \text{ is connected}: |V| = n, \\
& \qquad \qquad \qquad \qquad \qquad \qquad \; \; C_1 a_n \leq |E|
\leq C_2 a_n, \text{ and } C_1 b_n \leq \sum_{v \in V} \frac{1}{\sqrt{d_v}} \leq
C_2 b_n \Big \}.
\end{align*}
As noted in Section~\ref{sec:unperm}, the average-case design
observation model is equivalent to choosing the matrix $M^*$ from a
random ensemble with the permutation $\pi^*$ chosen uniformly at
random, and observing fixed pairwise comparisons. Such a viewpoint is
useful in order to state our lower bound. Expectations are taken over
the randomness of both $\pi^*$ and the Bernoulli observation noise.

\begin{theorem}
  \label{thm:lb1}
(a) Let $M^* = \Mns(\pi^*, 1/4)$, where the permutation $\pi^*$ is
  chosen uniformly at random on the set $[n]$. For any pair of sequences $\left( \{a_n\}_{n \geq 1}, \{b_n\}_{n \geq 1} \right)$
  such that the set $\mathcal{G}_n(a_n, b_n)$ is non-empty for every $n \geq 1$, and for any
  estimators $(\Mhat, \pihat)$ that are measurable functions of the
  observations on $G$, we have
\begin{align*}
\sup_{G \in \mathcal{G}_n(a_n, b_n)} \EE \left[ \frac{1}{n^2} \|\Mhat -
  M^* \|_F^2 \right] \gtrsim \frac{b_n}{n}, \text{ and } \sup_{G \in
  \mathcal{G}_n(a_n, b_n)} \EE \left[ \KT(\pi^*, \pihat) \right] \gtrsim
n b_n.
\end{align*}
(b) For any graph $G$, let $M^* = \Mns(\pi^*, c \sqrt{n / |E|})$, with
the permutation $\pi^*$ chosen uniformly at random and the constant
$c$ chosen sufficiently small. Then for any estimators $(\Mhat,
\pihat)$ that are measurable functions of the observations on $G$, we
have
\begin{align*}
\EE \left[ \frac{1}{n^2} \|\Mhat - M^* \|_F^2 \right] \gtrsim
\frac{n}{|E|}.
\end{align*}
\end{theorem}
Parts (a) and (b) of the lower bound may be interpreted respectively
as the approximation error caused by having observations only on a
subset of edges, and the estimation error arising from the Bernoulli
observation noise.  Note that part (b) applies to every graph, and is
particularly noteworthy for sparse graphs. In particular, in the
regime in which the graph has bounded average degree, it shows that
the inconsistency exhibited by the $\ASP$ estimator is unavoidable for any estimator. A more
detailed discussion for specific graphs may be found in
Section~\ref{sec:sim}.

Although part (a) of the theorem is stated for a supremum over graphs,
we actually prove a stronger result that explicitly characterizes the
class of graphs that attain these lower bounds. As an example, given
the sequences $a_n = n^2$ and $b_n = \sqrt{n}$, we show that the
$\ASP$ estimator is information-theoretically optimal for the sequence
of graphs consisting of two disjoint cliques $K_{n/2} \cup K_{n/2}$,
which can be verified to lie within the class $\mathcal{G}(a_n, b_n)$.

The $\ASP$ estimator for the SST model would replace step (iii), as
stated, by a maximum likelihood estimate using the entries on the
edges that we observe. However, analyzing such an estimator given only
a single sample on the entries $\Ospace$ is a challenging problem due
to dependencies between the different steps of the estimator, and the
difficulty of solving the associated matrix completion
problem. Consequently, we turn to an observation model consisting of
two random designs, and design a different estimator that renders the
matrix completion problem tractable.

%%%%%%%%%%%%%%%%%%%%%%%%%%%%%%%%%%%%%%%%%%%%%%%%%%%%%%%%%%%%%%%%%%%%%%%%%%%%%

\subsection{Two random designs: SST matrix estimation}
\label{sec:blockingestimator}

Recall the average-case setting with multiple random designs, as
described in Section~\ref{sec:unperm}, in which the comparison
topology is fixed ahead of time, but one can collect multiple
observations by assigning items to the vertices of the underlying
graph at random.  In this section, we rely on two such independent
observations $\Ospace_1$ and $\Ospace_2$ to design an estimator that
is consistent over the SST class.
%\mjwcomment{We need
%  to have better motivation for when this two design case would arise
%  in practice, apart from ``it is what allows us to prove our theorem,
%  because Ashwin lost the Pizza Bet, and Martin will collect upon it,
%  with interest of one slice per month, after Ashwin returns from
%  Microsoft.}  
In order to describe our estimator, we require some additional
notation. For any matrix $X \in [0,1]^{n\times n}$ such that $X +
X^\top = \mathbf{e}\, \mathbf{e}^\top$, we use \mbox{$\rspace(X) \defn
  X \mathbf{e}$} to denote the vector of its row sums.  Note that this
vector is related to the vector of scores, as defined in
equation~\eqref{EqnDefnScore}, via $\rspace(X) = (n-1) \tau(X) +
1/2$. 

Our estimator relies on the approximation of any matrix $M^* \in
\csst$ by a block-wise constant matrix, and we require some more
definitions to make this precise. For any vector $v \in
\mathbb{R}_+^n$, fix some value $t \in (0,n)$ and define a block
partition \mbox{$\mathsf{bl}_t(v)$} of $v$ as
\begin{align*}
  \left[ \mathsf{bl}_t(v) \right]_i = \left\{j \in [n] : v_{j} \in
  \big[\lfloor (i -1) t \rfloor \big], \lfloor i t \rfloor -1 ]
    \right\}.
\end{align*}
In particular, the blocking vector $\mathsf{bl}_t(\rspace(X))$
contains a partition of indices such that the row sums of the matrix
within each block of the partition are within a gap $t$ of each
other. Denote the set of all possible partitions of the set $[n]$ by
$\chi_n$. For any partition $C \in \chi_n$ of the indices $[n]$,
define the set of blocks $\mathcal{B}(C) = \{ S \times T : S, T \in
C\}$.

By definition, given a partition $C \in \chi_n$ of $[n]$, the set
$\mathcal{B}(C)$ is a partition of the set $[n] \times [n]$ into
blocks. We are now ready to describe the blocking operation.  For
indices $i,j \in [n]$, denote by $B_C(i,j)$ the block in
$\mathcal{B}(C)$ that contains the tuple $(i,j)$. Given a matrix $X
\in [0,1]^{n\times n}$ satisfying $X + X^\top = \mathbf{e}\,
\mathbf{e}^\top$, we define the blocked version of $X$ depending on
observations in a set $E \subseteq [n] \times [n]$ as
  \begin{align}
    \label{eq:projblock}
\left[ \Bl(X, C, E) \right]_{ij} =
\begin{cases}
\frac{1}{|B_C(i,j) \cap E|} \sum_{(k, \ell) \in B_C(i,j) \cap E}
X_{k\ell} &\text{ if } B_C(i,j) \cap E \neq \phi \\ 1/2 &\text{
  otherwise}.
\end{cases}
\end{align}
In words, this defines a projection of the matrix $X$ onto the
set of block-wise constant matrices, by block-wise averaging the
entries of $X$ over the observed set of entries $E$.  We now turn to
our estimator, called the Block-Average-Project estimator ($\BAP$ for short), of the underlying matrix
$M^* \in \csst$. Given the observation matrix $Y_1$, define
\begin{align*}
[Y'_1]_{ij} = 
\begin{cases}
\frac{n}{D_i} [Y_1]_{ij} &\text{ if entry }(i,j) \text{ is observed},
\\ 0 &\text{ otherwise,}
\end{cases}
\end{align*}
where $D_i = \sum_{j =1}^n [\Ospace_1]_{ij}$ is the (random) degree of
item $i$. We now perform three steps: \\ (1) {\bf Blocking step:} Fix
$S = \sum_{v \in V} 1/ \sqrt{d_v}$, and obtain the blocking vector
$\widehat{b} = \mathsf{bl}_S(\rspace(Y'_1))$ and permutation
$\pihatasp$ as in step (2) of the $\ASP$ estimator. \\ (2) {\bf
  Averaging step:} Average the matrix $Y_2$ within each block to
obtain the matrix $\widetilde{M} = \Bl(Y_2, \widehat{b}, E_2)$. \\ (3)
{\bf Projection step:} Project onto the space $\pihatasp(\cbiso) =
\{\pihatasp (M): M \in \cbiso\}$, to obtain the estimator $\Mhatbap$.

The blocking and averaging steps of the estimator are the main
ingredients that we use to bound the error of the associated matrix
completion problem. Also, the projection step of the estimator can be
computed in polynomial time via bivariate isotonic
regression~\cite{bril1984algorithm}.

\begin{theorem}
  \label{thm:sstupper}
Let the observation process be given by $\mathcal{O}_1 \cup
\mathcal{O}_2$. For any graph $G$ without isolated vertices and any
matrix $M^* \in \csst$, we have
\begin{align*}
\EE \left[ \frac{1}{n^2} \| \Mhatbap - M^* \|_F^2 \right] \lesssim
\frac{1}{n} \sum_{v \in V} \frac{1}{\sqrt{d_v}},
\end{align*}
where the expectation is taken over the noise, and observation
patterns $\Ospace_1$ and $\Ospace_2$.
\end{theorem}
To be clear, the blocking estimate $\Mhatbap$ is well-defined even
when we have just one sample $\mathcal{O}_1$ instead of two samples
$\mathcal{O}_1$ and $\mathcal{O}_2$, where step (2) is replaced by the
estimate $\widetilde{M} = \Bl(Y_1, \widehat{b}, E_1)$.  In the
simulations of Section~\ref{sec:sim}, we see that for a large variety
of graphs, using a single sample $\Ospace_1$ enjoys similar
performance to using two independent samples $\Ospace_1$ and
$\Ospace_2$.  We require two independent samples of the observations
in our theoretical analysis to decouple the randomness of the first
step of the algorithm from the second.  When using one sample
$\Ospace_1$, the dependencies that are introduced between the
different steps of the algorithm make the analysis challenging.

% Local Variables:
% TeX-master: "worstavg_design_arxiv"
% End: 

\section{Dependence on graph topologies}
\label{sec:sim}

In this section, we discuss implications of our results for some
comparison topologies. Let us focus first on the worst-case design
setting, and the lower bound of Theorem~\ref{thm:minimax}.  For the
star, path (or more generally, any graph with bounded average degree),
and complete bipartite graphs, one can verify that we have $\alpha(G)
\asymp n$, so $\Mrisk(G, \Cns) \asymp 1$. If the graph is a union of
disjoint cliques $K_{n/2} \cup K_{n/2}$ (or having a constant number
of edges across the cliques, like a barbell graph), then we see that
$\beta(G^c) \asymp n^2$, so $\Mrisk(G, \Cns) \asymp 1$.  Thus, our
theory yields pessimistic results for many practically motivated
comparison topologies under worst-case designs, even though all the
connected graphs above admit consistent estimation for parametric
models\footnote{The complete bipartite graph, for instance, admits
  optimal rates of estimation.} as the number of samples grows.  In
the average case-setting of Section~\ref{sec:unperm},
Theorems~\ref{thm:ubeff}, \ref{thm:lb1} and~\ref{thm:sstupper}
characterize the mean-squared Frobenius norm errors of the
corresponding estimators (up to constants) as \mbox{$\mathcal{D}(G)
  \defn \frac{1}{n} \sum_{v \in V} \frac{1}{\sqrt{d_v}}$.}

In order to illustrate our results for the average-case setting, we
present the results of simulations on data generated
synthetically\footnote{Note that the SST model has been validated
  extensively on real data in past work (see, e.g. Ballinger and
  Wilcox~\cite{ballinger1997decisions}).} from two special cases of
the SST model.  We fix $\pi^* = \id$ without loss of generality, and
generate the ground truth comparison matrix $M^*$ in one of two ways:
\begin{enumerate}
  \item[(1)] Noisy sorting with high SNR: We set $M^* = \Mns(\id,
    0.4)$.
\item[(2)] SST with independent bands: We first set $M^*_{ii} = 1/2$
  for every $i$.  Entries on the diagonal band immediately above the
  diagonal (i.e. $M^*_{i,i+1}$ for $i \in [n-1]$) are chosen
  i.i.d. and uniformly at random from the set $[1/2,1]$. The band
  above is then chosen uniformly at random from the allowable set,
  where every entry is constrained to be upper bounded by $1$ and
  lower bounded by the entries to its left and below. We also set
  $M^*_{ij} = 1 - M^*_{ji}$ to fill the rest of the matrix.
\end{enumerate}

For each graph $G$ with adjacency matrix $A$, the data is generated
from ground truth by observing independent Bernoulli comparisons under
the observation process $\Ospace = \sigma(A)$, for a randomly
generated permutation $\sigma$. For the SST model, we also generate
data from two independent random observations $\Ospace_1$ and
$\Ospace_2$ as required by the $\BAP$ estimator; however, we also
simulate the behaviour of the estimator for one sample $\Ospace_1$ and
show that it closely tracks that of the two-sample estimator.

\begin{figure}
    \centering
    \begin{subfigure}{0.48\textwidth}
        \centering
        \includegraphics[scale=0.38]{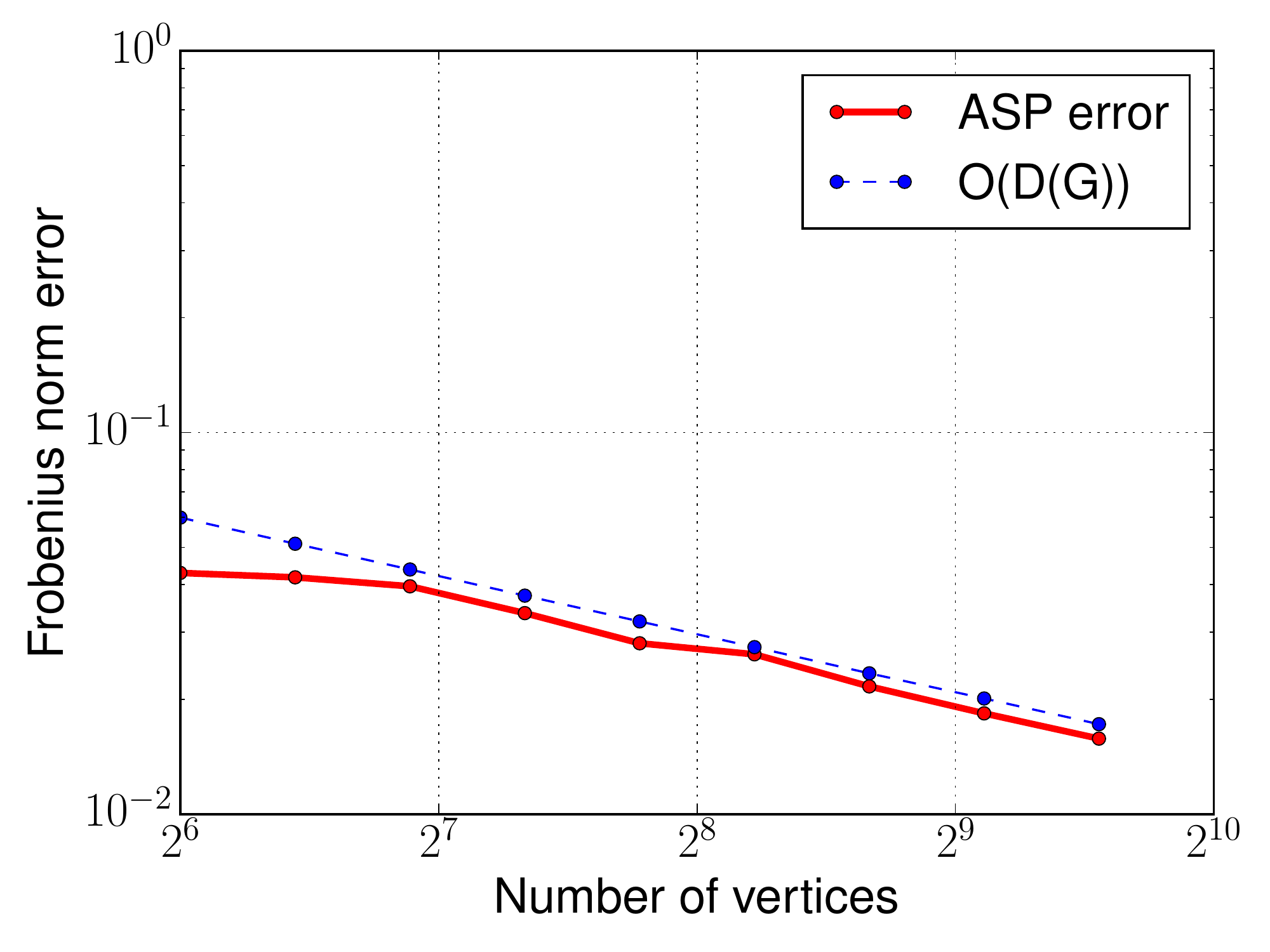}
        \caption{Two-disjoint-cliques}
    \end{subfigure}
    %\begin{subfigure}{0.3\textwidth} \centering
    %    \includegraphics[scale=0.25]{Figures/bipartiteNS.png} \caption{Complete
    %    bipartite graph.}  \end{subfigure}
    \begin{subfigure}{0.48\textwidth}
        \centering
        \includegraphics[scale=0.38]{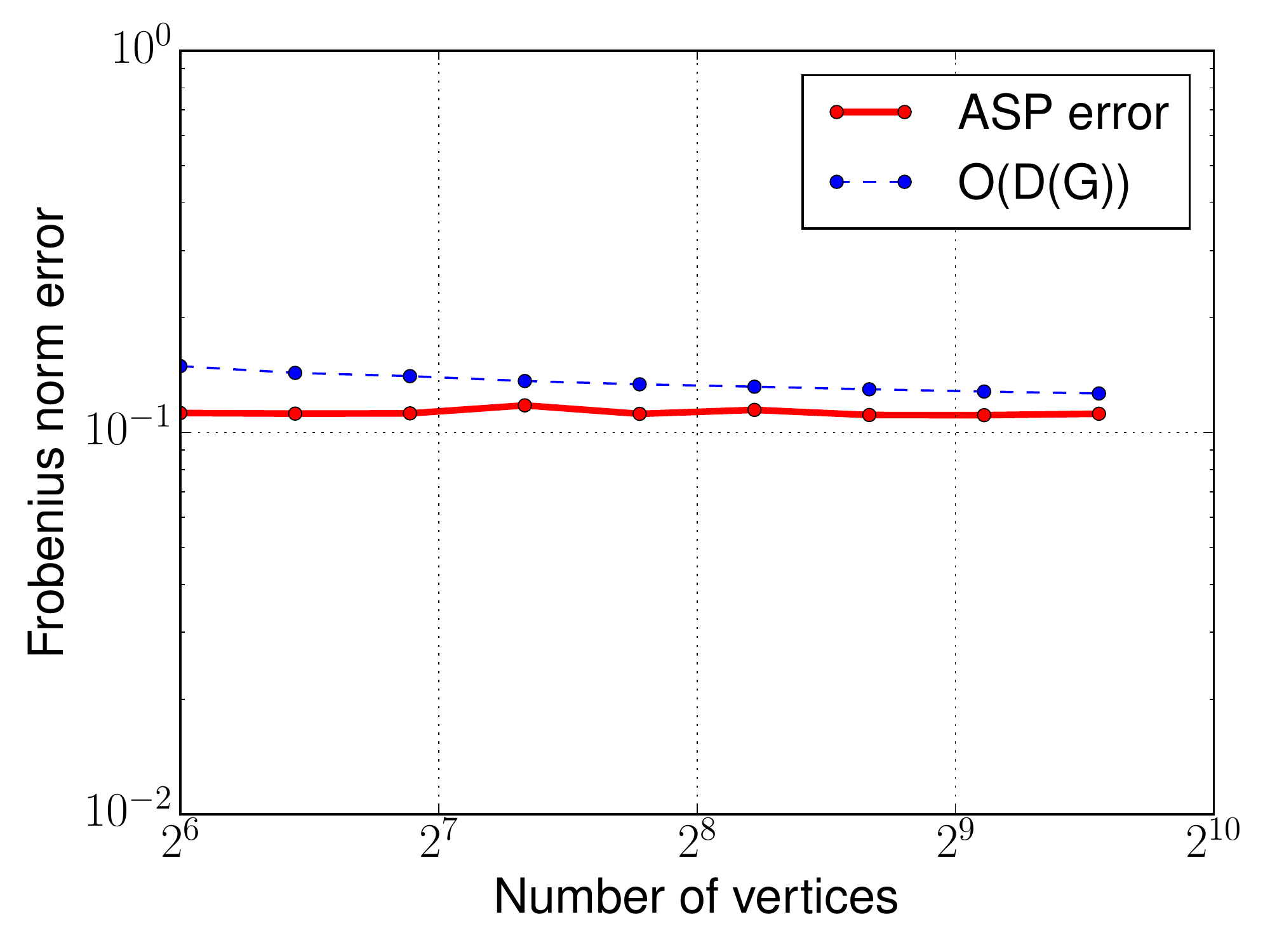}
        \caption{Clique-plus-path}
    \end{subfigure}
    \begin{subfigure}{0.48\textwidth}
        \centering
        \includegraphics[scale=0.38]{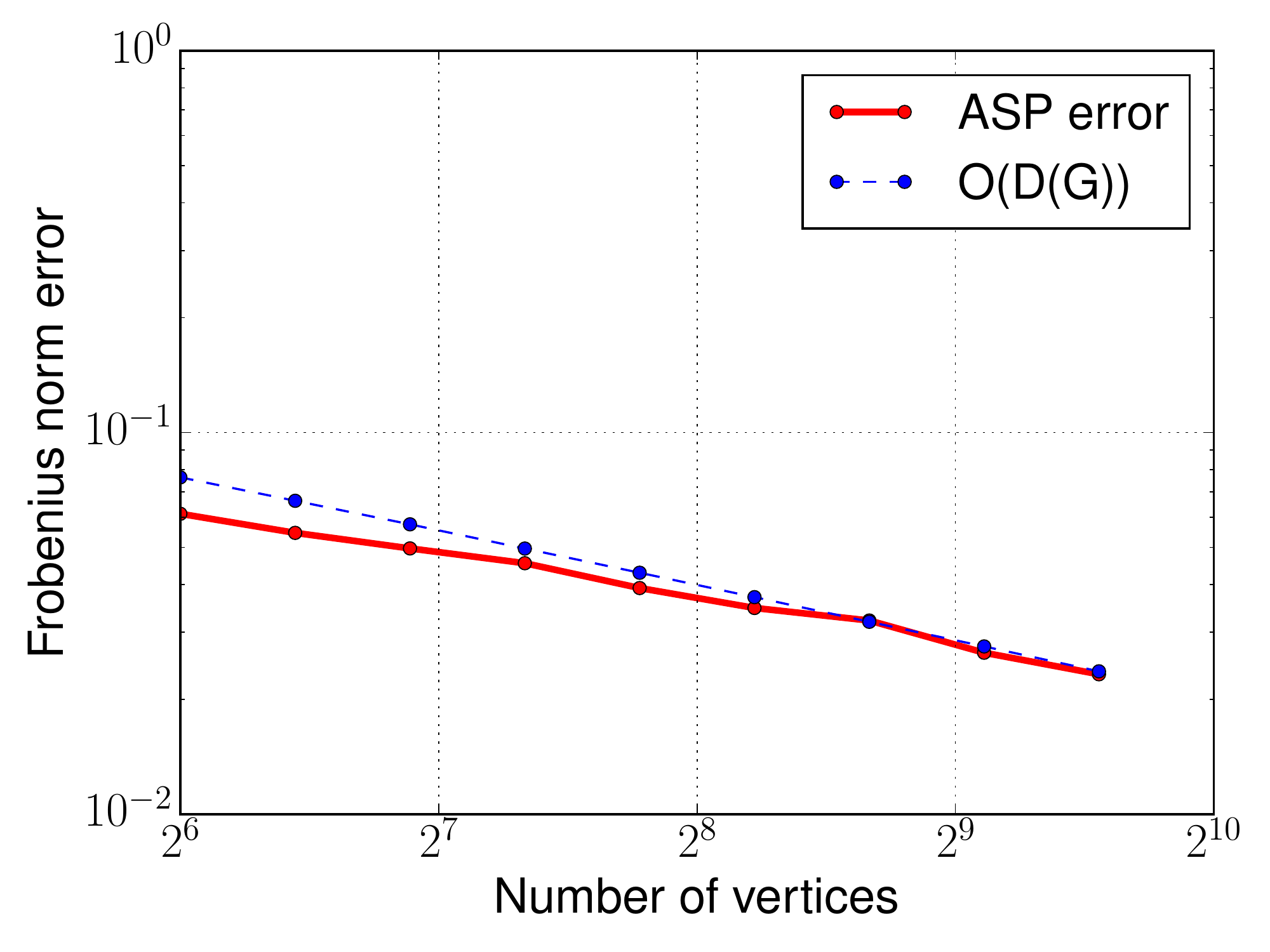}
        \caption{Power law with $d_i = i$}
    \end{subfigure}
    %\begin{subfigure}{0.24\textwidth} \centering
    %    \includegraphics[scale=0.18]{Figures/SBMNS.png} \caption{Stochastic
    %    block model graph with $p = 0.9$, $q = 0.1 \frac{\log
    %    n}{n}$.}  \end{subfigure}
    \begin{subfigure}{0.48\textwidth}
    	\centering
    	\includegraphics[scale=0.38]{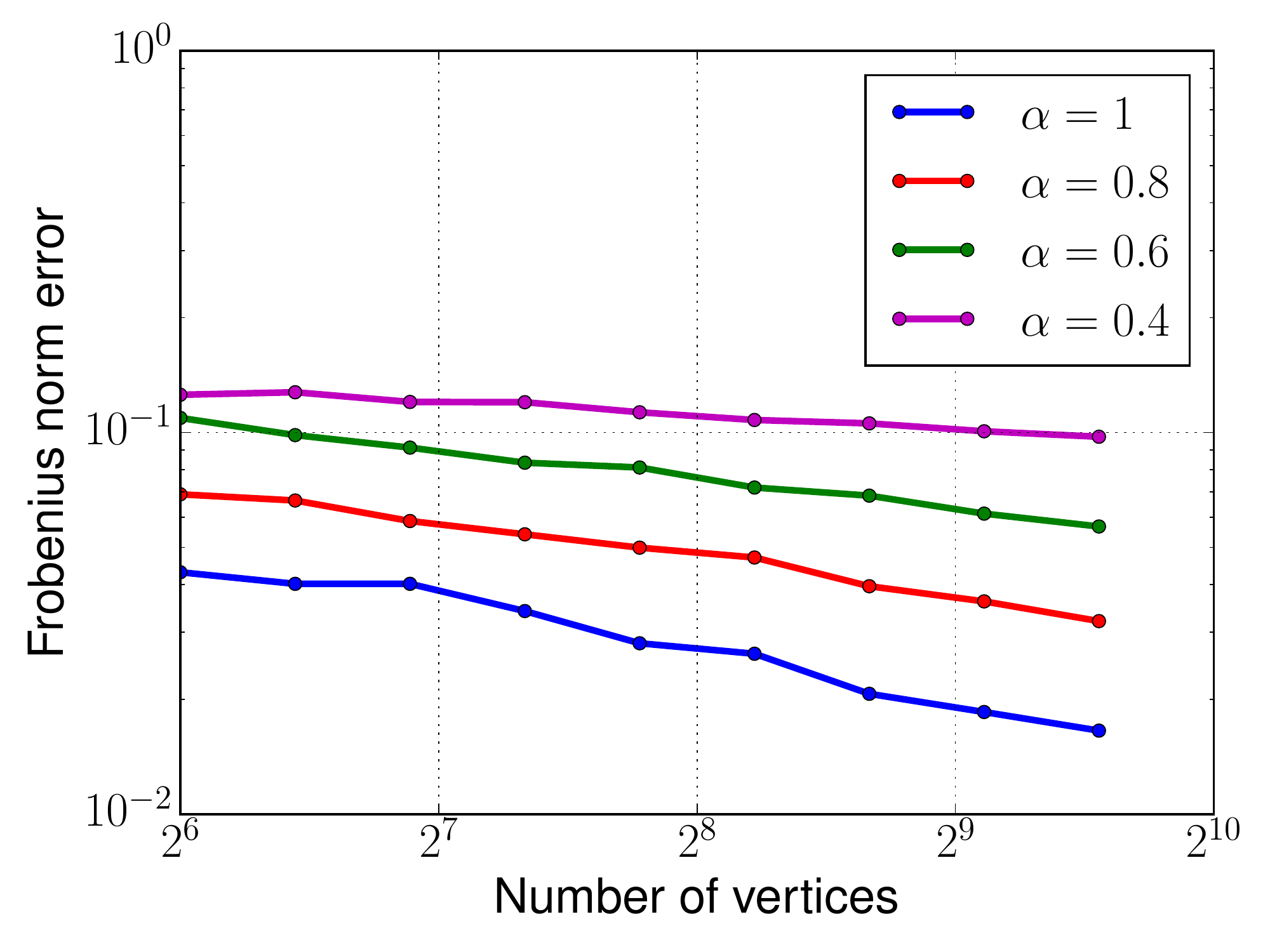}
    	\caption{$\lfloor (\frac n2)^\alpha \rfloor$-regular bipartite}
    \end{subfigure}
    \caption{Normalized Frobenius norm error $\frac{1}{n^2} \|
      \Mhatasp - M^*\|_F^2$ with data generated using the noisy
      sorting model $M^* = \Mns(\id, 0.4)$, averaged over $10$
      trials.} \label{fig:nsplots}
\end{figure}

\begin{figure}
    \centering
    \begin{subfigure}{0.48\textwidth}
        \centering
       \includegraphics[scale=0.38]{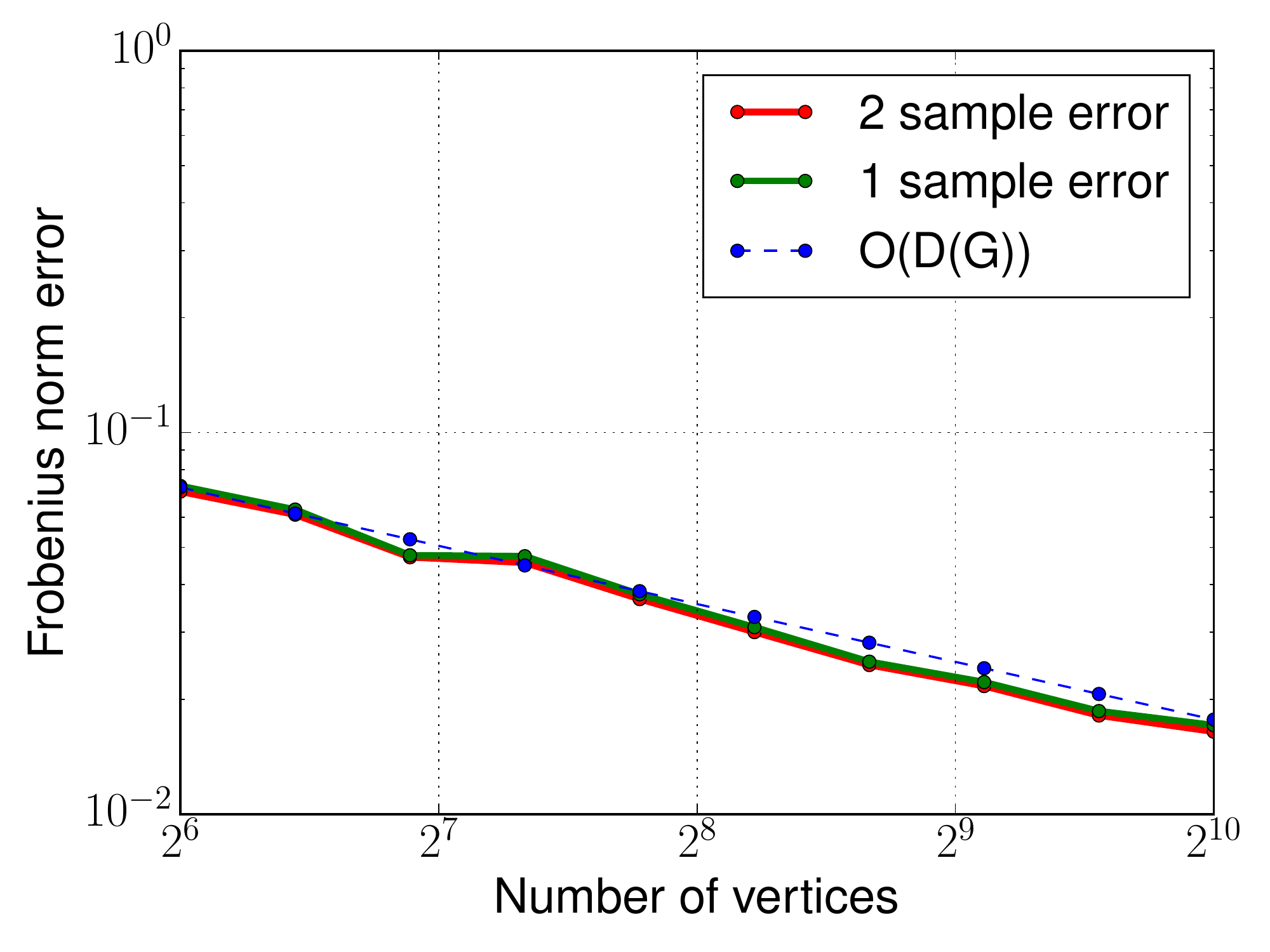}
        \caption{Two-disjoint-cliques}
    \end{subfigure} 
    %\begin{subfigure}{0.3\textwidth} \centering
    %    \includegraphics[scale=0.25]{Figures/bipartiteSSTIndependentBands_new.png} \caption{Complete
    %    bipartite graph.}  \end{subfigure}
    \begin{subfigure}{0.48\textwidth}
        \centering
        \includegraphics[scale=0.38]{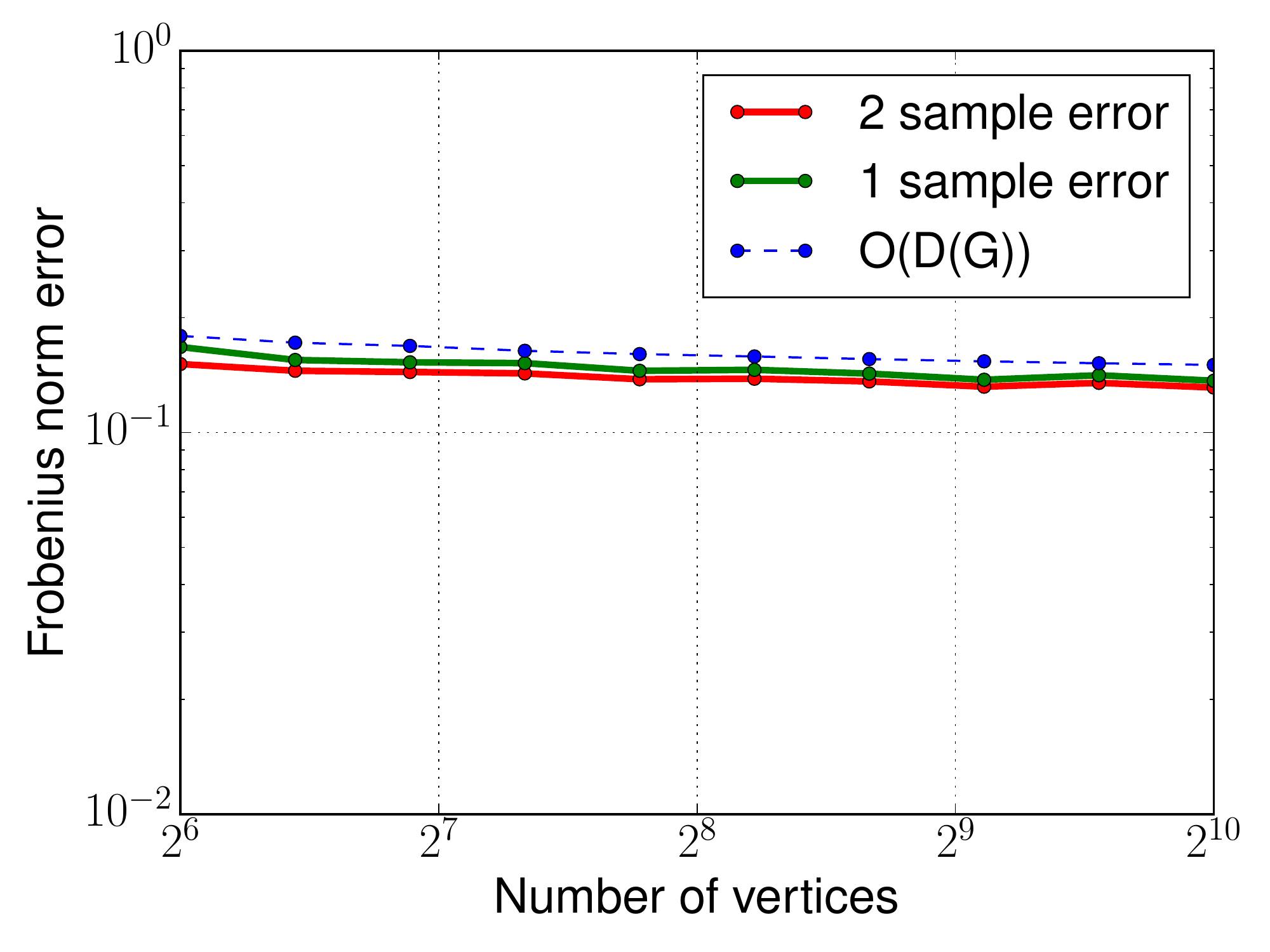}
        \caption{Clique-plus-path}
    \end{subfigure}
    \begin{subfigure}{0.48\textwidth}
        \centering
        \includegraphics[scale=0.38]{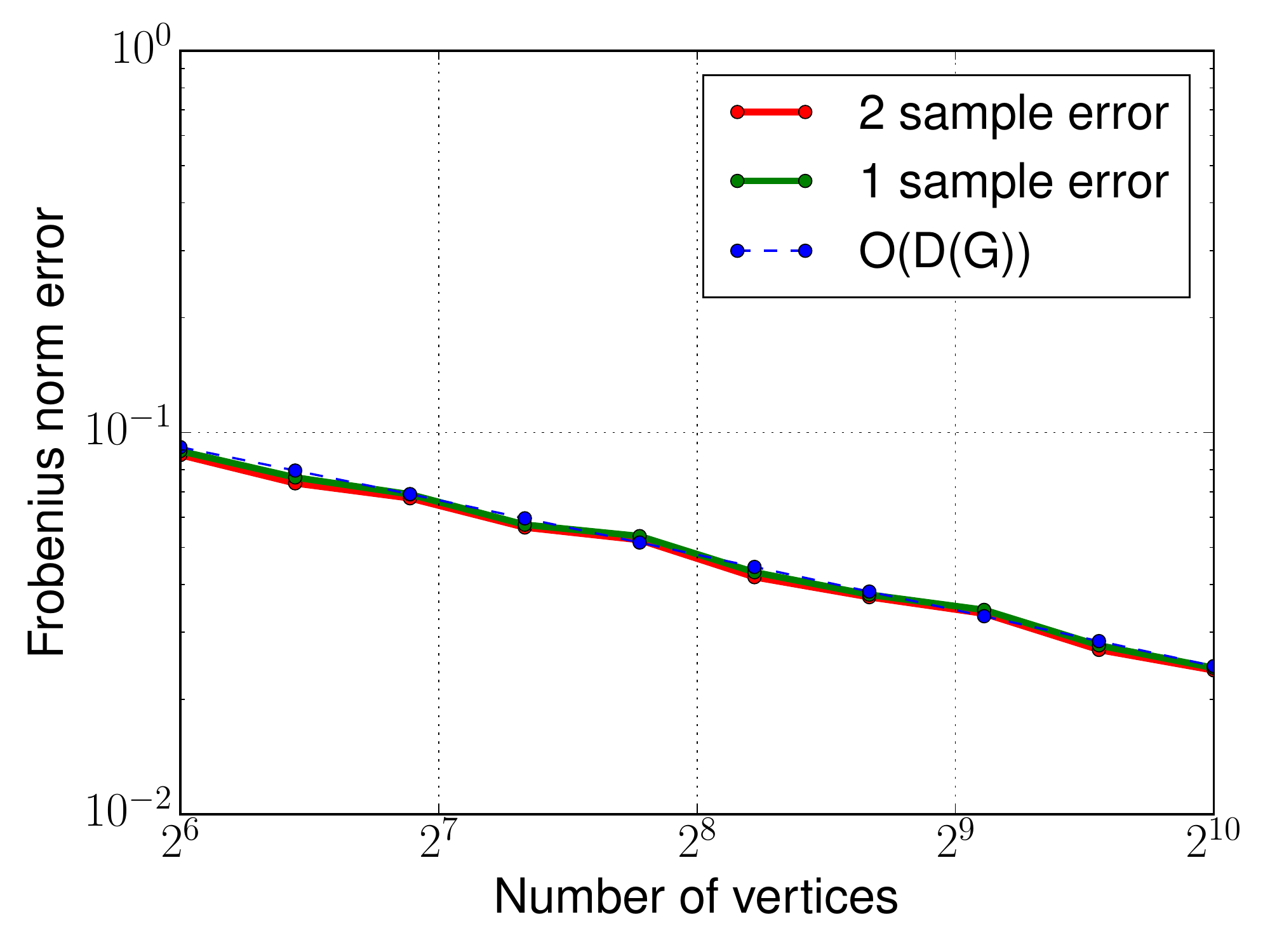}
        \caption{Power law with $d_i = i$}
    \end{subfigure}
    %\begin{subfigure}{0.24\textwidth}
    %    \centering
    %    \includegraphics[scale=0.18]{Figures/SBMSSTIndependentBands.png}
    %    \caption{Stochastic block model graph with $p = 0.9$, $q = 0.1 \frac{\log n}{n}$.}
    %\end{subfigure}
    \begin{subfigure}{0.48\textwidth}
        \centering
        \includegraphics[scale=0.38]{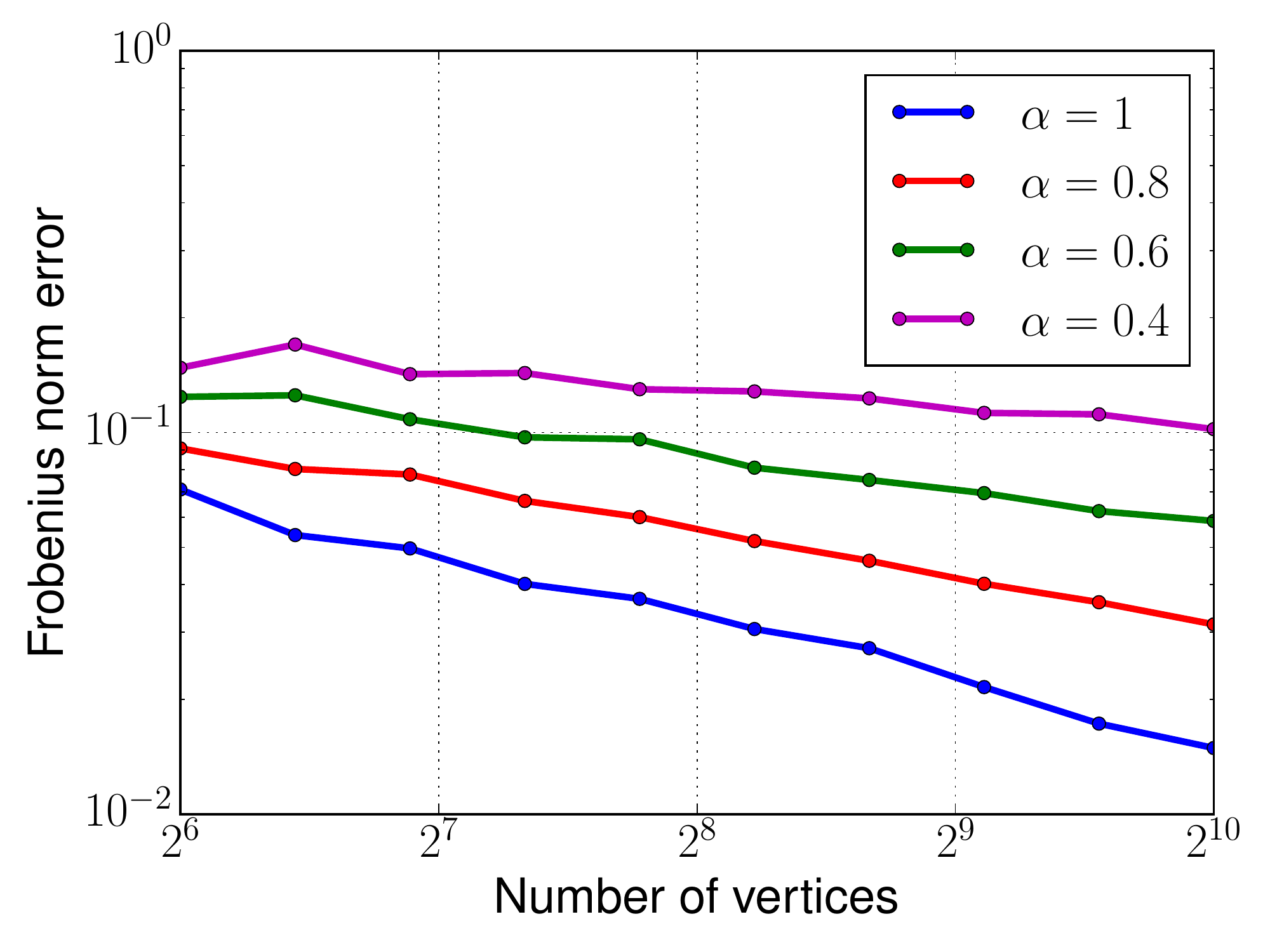}
        \caption{$\lfloor (\frac n2)^\alpha \rfloor$-regular bipartite}
    \end{subfigure}
    \caption{Normalized Frobenius norm error $\frac{1}{n^2} \|
      \Mhatbap - M^*\|_F^2$ with data generated using the SST model
      with independent bands, averaged over $10$ trials, plotted for
      one and two samples.} \label{fig:sstplots}
\end{figure}

Recall that the estimation error rate was dictated by the degree
functional $\mathcal{D}(G)$. While our graphs were chosen to
illustrate scalings of $\mathcal{D}(G)$, some variants of these graphs
also naturally arise as comparison topologies.\\
(1) {\bf Two-disjoint-clique graph:} For this graph $K_{n/2} \cup
K_{n/2}$, we have $d_v = \frac{n}{2} - 1$ for every $v \in V$, and
simple calculations yield $\mathcal{D}(G) \asymp \frac{1}{\sqrt{n}}$.
It is interesting to note that this graph has unfavorable guarantees
for parametric estimation under the adversarial model, because it is
disconnected (and thus has a Laplacian with zero spectral gap.) We
observe that this spectral property does not play a role in our
analysis of the $\ASP$ or $\BAP$ estimator under the average-case
observation model, and this behavior is corroborated by our
simulations. Although we do not show it here, a similar behavior is
observed for the stochastic block model, a practically motivated
comparison topology when there are genres present among the items,
which is a relaxation of the two-clique case allowing for sparser
``communities" instead of cliques, and edges between the
communities. \\
(2) {\bf Clique-plus-path graph:} The nodes are partitioned into two
sets of $n/2$ nodes each.  The graph contains an edge between every
two nodes in the first set, and a path starting from one of the nodes
in the first set and chaining the other $n/2$ nodes.  This is an
example of a graph construction that has many ($\asymp n^2$) edges,
but is unfavorable for noisy sorting or SST estimation.  Simple
calculations show that the degree functional is dominated by the
constant degree terms and we obtain $\mathcal{D}(G) \asymp 1$. \\
(3) {\bf Power law graph:} We consider the special power law
graph~\cite{barabasi1999emergence} with degree sequence $d_i = i$ for
$1 \leq i \leq n$, and construct it using the Havel-Hakimi
algorithm~\cite{havel1955remark, hakimi1962realizability}.  For this
graph, we have a disparate degree sequence, but $\mathcal{D}(G) \asymp
\frac{1}{\sqrt{n}}$, and the simulated estimators are consistent. \\
(4) $\lfloor (n/2)^{\alpha} \rfloor$-{\bf regular bipartite graphs:} A
final powerful illustration of our theoretical guarantees is provided
by a regular bipartite graph construction in which the nodes are
partitioned into two sets of $n/2$ nodes each, and each node in one
set is (deterministically) connected to $\lfloor (n / 2)^{\alpha}
\rfloor$ nodes in the other set.  This results in the degree sequence
$d_v = \lfloor (n /2)^{\alpha} \rfloor$ for all $v \in V$, and the
degree functional evaluates to $\mathcal{D}(G) \asymp n^{-\alpha/2}$.
The value of $\alpha$ thus determines the scaling of the estimation
error for the $\ASP$ estimator in the noisy sorting case, as well as
the $\BAP$ estimator in the SST case, as seen from the slopes of the
corresponding plots.

Some other graphs that were considered in parametric model
environments~\cite{shah16estimation}, such as the star, cycle, path
and hypercube graphs, turn out to be unfavorable for permutation-based
models even in the average-case setting, as corroborated by the lower
bound of Theorem~\ref{thm:lb1}, part (b).

% Local Variables:
% TeX-master: "worstavg_design_arxiv"
% End: 

\section{Proofs}
\label{sec:proof}

In this section, we provide the proofs of our main results. We assume
throughout that $n \geq 2$, and use $c, c'$ to denote universal
constants that may change from line to line.

%%%%%%%%%%%%%%%%%%%%%%%%%%%%%%%%%%%%%%%%%%%%%%%%%%%%%%%%%%%%%%%%%%%%%%%%%%%%%%%%

\subsection{Proof of Theorem~\ref{thm:minimax}}

For each fixed graph $G$, define the quantity
\begin{align*}
\Arisk(G) \defn \sup_{\substack{M, M' \in \Cns \\ M(G) = M'(G)}}
\frac{1}{n^2} \sum_{(i,j) \notin E} (M_{ij} - M'_{ij} )^2
\end{align*}
corresponding to the diameter quantity that is lower bounded in
equation~\eqref{eq:dia}.  Taking the lower
bound~\eqref{eq:dia} as given for the moment, we first prove the lower
bound~\eqref{eq:minimax} on the minimax risk.  It suffices to show that the minimax risk is lower bounded in
terms of $\Arisk(G)$ as
\begin{align}
  \label{eq:approxlb}
\inf_{\Mhat = f(Y(G))} \sup_{M^* \in \Cns} \EE \left[ \frac{1}{n^2}
  \|\Mhat - M^* \|_F^2 \right] \geq \frac{1}{4} \Arisk(G).
\end{align}
In order to verify this claim, consider the two matrices $M^1, M^2 \in
\csst$ that attain the supremum in the definition of $\Arisk(G)$; note
that such matrices exist due to the compactness of the space and the
continuity of the squared loss.  By construction, these two matrices
satisfy the properties
\begin{align*}
M^1(G) = M^2(G), \quad \text{ and } \quad \sum_{(i, j) \notin E}
(M^1_{ij} - M^2_{ij})^2 = n^2 \Arisk(G).
\end{align*}

We can now reduce the problem to one of testing between the two matrices
$M^1$ and $M^2$, with the distribution of observations being identical 
for both alternatives. Consequently, any procedure can do no better than
to make a random guess between the two, so we have
%Now suppose that we reveal $M^1(G)$ to the procedure. Note that a
%lower bound with such side information available constitutes a lower
%bound for the original problem, since any procedure can simply ignore
%the side information, if it so chooses. By construction, we have ensured
%that the distribution of the observation

%Suppose that the procedure
%outputs the matrix $\Mhat$ as an estimate, and set
%\begin{align*}
%M^* = \arg\max_{M^k \in \{M^1, M^2\}} \sum_{(i,j) \notin E}
%(\Mhat_{ij} - M^k_{ij} )^2.
%\end{align*} 
%By our construction combined with the triangle inequality, we have
\begin{align*}
\inf_{\Mhat} \sup_{M^* \in \Cns} \EE \left[\|\Mhat - M^* \|_F^2 \right] \geq \frac{1}{4} \sum_{(i,j) \notin E} (M^1_{ij} -
M^2_{ij})^2,
\end{align*}
  which proves the claim~\eqref{eq:approxlb}.

It remains to prove the claimed lower bound~\eqref{eq:dia} on
$\Arisk(G)$.  This lower bound can be split into the following two
claims:
\begin{subequations}
\begin{align}
\Arisk(G) &\geq \frac{1}{n^2} \alpha(G) (\alpha(G) - 1), \text{
  and}\label{eq:approx1}\\ \Arisk(G) &\geq \frac{1}{n^2}
\beta(G^c). \label{eq:approx2}
\end{align}
\end{subequations}
We use a different argument to establish each claim.\\

%%%%%%%%%%%%%%%%%%%%%%%%%%%%%%%%%%%%%%%%%%%%%%%%%%%%%%%%%%%%%%%%%%%%%%%%%%%%%%%%%%
\noindent\textbf{Proof of claim~\eqref{eq:approx1}:} Recall from
Section~\ref{sec:intro} the definition of the largest independent set.
Without loss of generality, let the largest independent set be given
by $I = \{v_1, \ldots v_\alpha\}$. Assign item $i$ to vertex $v_i$ for
$i \in [\alpha]$. Now we choose permutations $\pi$ and $\pi'$ so that
\begin{itemize}
\item $\pi(i) = i$ for $i \in [\alpha]$,
\item $\pi'(i) = \alpha-i+1$ for $i \in [\alpha]$,
\item $\pi$ and $\pi'$ agree on $\{\alpha+1, \dots, n\}$.
\end{itemize}
Note that last step is possible because $\pi([\alpha]) =
\pi'([\alpha])$. Moreover, define the matrices $M = \Mns(\pi, 1/2)$
and $M' = \Mns(\pi',1/2)$.  Note that by construction, we have ensured
that $M(G) = M'(G)$. However, it holds that
\begin{align*}
\sum_{(i,j) \notin E} (M_{ij} - M'_{ij} )^2 = \| M - M' \|_F^2 = 2 \,
\KT(\pi, \pi') = \alpha(\alpha-1),
\end{align*}
which completes the proof. \\

\smallskip
\noindent\textbf{Proof of claim~\eqref{eq:approx2}:} Recall the
definition of a maximum biclique from Section~\ref{sec:intro}. Since
the complement graph $G^c$ has a biclique with $\beta(G^c)$ edges, the
graph $G$ has two disjoint sets of vertices $V_1$ and $V_2$ with
$|V_1||V_2| = \beta(G^c)$ that do not have edges connecting one to the
other. We now pick the two permutations $\pi$ and $\pi'$ so that
\begin{itemize}
\item the permutation $\pi$ ranks items from $V_1$ as the top $|V_1|$
  items, and ranks items from $V_2$ as the next $|V_2|$ items;
\item the permutation $\pi'$ ranks items from $V_2$ as the top $|V_2|$
  items, and ranks items from $V_1$ as the next $|V_2|$ items;
\item the permutations $\pi$ and $\pi'$ agree with each other apart
  from the above constraints.
\end{itemize}
As before, we define $M = \Mns(\pi,1/2)$ and $M' = \Mns(\pi',1/2)$,
and again, we have \mbox{$M(G) = M'(G)$.} The relative orders of items
have been interchanged across the biclique, so it holds that $2\,
\KT(\pi, \pi') = \beta(G^c)$, which completes the proof.  \qed

%%%%%%%%%%%%%%%%%%%%%%%%%%%%%%%%%%%%%%%%%%%%%%%%%%%%%%%%%%%%%%%%%%%%%%%%%%%%%

\subsection{Some useful lemmas for average-case proofs}

We now turn to proofs for the average-case setting.  For convenience,
we begin by stating two lemmas that are used in multiple proofs.  The
first lemma bounds the performance of the permutation estimator
$\pihatasp$ for a general SST matrix, and is thus of independent
interest.
\begin{lemma}
\label{lem:scores}
For any matrix $M^* \in \csst$, the permutation estimator $\pihatasp$
satisfies
\begin{subequations}
\begin{align} \label{eq:mtau1}
\| \pihatasp(M^*) - M^* \|_F^2 \leq 4 (n - 1) \| \tau^* - \tauhat
\|_1,
\end{align}
and if additionally, $M^* \in \Cns(\lambda^*)$, we have
\begin{align}
\label{eq:mtau2}
\| \pihatasp(M^*) - M^* \|_F^2 \leq 8 \lambda^* (n-1) \| \tau^* -
\tauhat \|_1.
\end{align}
\end{subequations}
In addition, the score estimates satisfy the bounds
\begin{align*}
\EE [\| \tau^* - \tauhat \|_1] \leq c \sum_{v \in V}
\frac{1}{\sqrt{d_v}}, \quad \text{ and } \quad \Pr \Big\{\| \tau^* -
\tauhat \|_1 \geq c \sqrt{\log n} \sum_{v \in V} \frac{1}{\sqrt{d_v}}
\Big\} \leq n^{-10}.
\end{align*}
\end{lemma}
\noindent Note that Lemma~\ref{lem:scores} implies the
bound~\eqref{eq:nsubkt}, since for a matrix $M^* \in \Cns(\lambda^*)$,
we have $8 \lambda^2 \KT(\pihatasp, \pi^*) = \| \pihatasp(M^*) - M^*
\|_F^2$.\\

\noindent Our second lemma is a type of rearrangement inequality.
\begin{lemma}
\label{lem:rearrange}
Let $\{a_u\}_{u=1}^n$ be an increasing sequence of positive numbers
and let $\{b_u\}_{u=1}^n$ be a decreasing sequence of positive
numbers. Then we have
\begin{align*}
\Big(\sum_{u=1}^n a_u\Big) \Big(\sum_{u=1}^n b_u\Big) \ge n
\sum_{u=1}^n a_u b_u.
\end{align*}
\end{lemma}

%%%%%%%%%%%%%%%%%%%%%%%%%%%%%%%%%%%%%%%%%%%%%%%%%%%%%%%%%%%%%%%%%%%%%%%%%%%
\subsubsection{Proof of Lemma~\ref{lem:scores}}

Assume without loss of generality that $\pi^* = \id$. We begin by
applying H\"older's inequality to obtain
\begin{align*}
\| \pihatasp(M^*) - M^* \|_F^2 & \leq \|\pihatasp(M^*) - M^* \|_\infty
\|\pihatasp(M^*) - M^* \|_1.
\end{align*}
In the case where $M^* \in \Cns(\lambda^*)$, we have $\|
M^*_{\pihatasp(i)} - M^*_i \|_\infty \leq 2\lambda^*$; in the general
case $M^* \in \csst$, we have $\| M^*_{\pihatasp(i)} - M^*_i \|_\infty
\leq 1$. Next, if $M^*_{\pihatasp}$ denotes the matrix obtained from
permuting the rows of $M^*$ by $\pihatasp$, then it holds that
\begin{align*}
\|\pihatasp(M^*) - M^* \|_1 &\leq \|\pihatasp(M^*) - M^*_{\pihatasp}
\|_1 + \|M^*_{\pihatasp} - M^* \|_1\\ &= 2 \sum_{i=1}^n \|
M^*_{\pihatasp(i)} - M^*_i \|_1,
\end{align*}
where the equality follows from the condition $M^*_{ij}+M^*_{ji}=1$.
We also have
\begin{align*}
\sum_{i=1}^n \| M^*_{\pihatasp(i)} - M^*_i \|_1 &\stackrel{\1}{=}
(n-1) \sum_{i=1}^n | \tau^*_{\pihatasp(i)} - \tau^*_i | \notag \\
& = (n-1) \sum_{i=1}^n \big| \tau^*_{i} - \tau^*_{\pihatasp^{-1}(i)}
\big| \notag\\
& \leq (n-1) \left[ \sum_{i=1}^n \big| \tau^*_i -
  \tauhat_{\pihatasp^{-1}(i)} \big| + \sum_{i=1}^n \big|
  \tauhat_{\pihatasp^{-1}(i)} - \tau^*_{\pihatasp^{-1}(i)} \big|
  \right] \\
& \stackrel{\2}{\leq} (n-1) \left[ \sum_{i=1}^n |
  \tau^*_{i} - \tauhat_{i} | + \sum_{i=1}^n | \tauhat_i - \tau^*_i |
  \right] \\
& = 2 (n-1) \| \tau^* - \tauhat \|_1,
\end{align*}
where step $\1$ is due to monotonicity along each column of $M^*$, and
step $\2$ follows from the $\ell_1$-rearrangement inequality (see,
e.g., Example~2 in the paper~\cite{vince90a}), using the fact that
both sequences $\{\tau^*_i\}_{i=1}^n$ and
$\{\tauhat_{\pihatasp^{-1}(i)}\}_{i=1}^n$ are sorted in decreasing
order. Combining the last three displays yields the claimed
bounds~\eqref{eq:mtau1} and~\eqref{eq:mtau2}.

In order to prove the second part of the lemma, it suffices to show
that the random variable $\| \tau^* - \tauhat \|_1$ is sub-Gaussian
with parameter $cS$, where $S \defn \sum_{v\in V} 1/\sqrt{d_v}$. Let
$\sigma:[n] \to V$ be the uniform random assignment of items to
vertices with $\sigma(A) = \Ospace$, and let $D_i$ denote the random
degree $d_{\sigma(i)} = \sum_{j \neq i} \Ospace_{ij}$ of item $i$.
Note that conditioned on the event $\sigma(i) = v$, the difference
between a score and its empirical version can be written as
\begin{align*}
  \hat \tau_i - \tau^*_i = \Big(\frac{1}{d_v} \sum_{j: \sigma(j) \sim
    v} M^*_{ij} - \frac{1}{n-1} \sum_{j\neq i} M^*_{ij}\Big) +
  \frac{1}{d_v} \sum_{j: \sigma(j) \sim v} W_{ij},
\end{align*}
where $\sim$ denotes the presence of an edge between two vertices.
The term $\frac{1}{d_v} \sum_{j: \sigma(j) \sim v} M^*_{ij}$ is the
empirical mean of $d_v$ numbers chosen uniformly at random without
replacement from the set $\{M^*_{ij}\}_{j \neq i}$, while
$\frac{1}{n-1} \sum_{j\neq i} M^*_{ij}$ is the true
expectation. Moreover, $W_{ij}$ represents independent, zero-mean
noise bounded within the interval $[-1,1]$. Consequently, applying
Hoeffding's inequality for sampling without
replacement~\cite[Proposition 1.2]{bardenet15concentration} and the
standard Hoeffding bound~\cite{hoeffding1963probability} to the two
parts respectively, we obtain
\begin{align} \label{eq:hoeffding} 
\Pr \big\{ | \hat \tau_i - \tau^*_i | \ge t \mid \sigma(i) = v \big\}
\le 4 \exp ( - c \, d_v t^2 ).
\end{align}
Replacing $t$ by $t/\sqrt{d_v}$, we see that conditioned on the event
$\sigma(i)=v$, the random variable $\sqrt{d_v} | \hat \tau_i -
\tau^*_i |$ is sub-Gaussian with a constant parameter $c'$, or
equivalently,
\begin{align} \label{eq:cond-subg} 
\E \Big[ \exp \Big( t \sqrt{D_i} \, | \hat \tau_i - \tau^*_i | \Big) \mid \sigma(i) = v \Big] \le \exp (c \, t^2) \,. 
\end{align}
Since $S = \sum_{i=1}^n 1/\sqrt{D_i}$, Jensen's inequality implies that
\begin{align} 
\E \Big[ \exp \Big( t \sum_{i=1}^n | \hat \tau_i - \tau^*_i | \Big)
  \Big]
& \le \E \Big[ \sum_{i=1}^n \frac{1/\sqrt{D_i}}{S} \exp \Big( t S \sqrt{D_i} \, | \hat \tau_i - \tau^*_i | \Big) \Big] \notag \\
& = \sum_{i=1}^n \frac{1}{S} \sum_{v \in V} \Pr \big\{ \sigma(i) = v
\big\} \E \Big[ \frac{1}{\sqrt{D_i}} \exp \Big( t S \sqrt{D_i} \, |
  \hat \tau_i - \tau^*_i | \Big) \mid \sigma(i) = v\Big] \notag \\
& \le \sum_{i=1}^n \frac{1}{S} \sum_{v \in V} \frac{1}{n}
\frac{1}{\sqrt{d_v}} \exp(c S^2 t^2) \notag \\
& = \exp(c S^2 t^2) \,, \notag
\end{align}
where the last inequality follows from
equation~\eqref{eq:cond-subg}. Therefore, the random variable $\| \hat
\tau - \tau^* \|_1$ is sub-Gaussian with parameter $c S$, as claimed.
\qed

%%%%%%%%%%%%%%%%%%%%%%%%%%%%%%%%%%%%%%%%%%%%%%%%%%%%%%%%%%%%%%%%%%%%%%%%%
\subsubsection{Proof of Lemma~\ref{lem:rearrange}}

For any increasing sequence $\{a_u\}$ and decreasing sequence
$\{b_u\}$, the rearrangement inequality (see, e.g., Example~2 in the
paper~\cite{vince90a}) guarantees that
\begin{align*}
  \sum_{u =1}^n a_u b_u \leq \sum_{u = 1}^n a_u b_{\pi(u)} \qquad
  \mbox{for any permutation $\pi$.}
\end{align*}
This inequality implies that
\begin{align*}
\frac{1}{n}(\sum_{u=1}^n a_u)(\sum_{u=1}^n b_u) = \frac{1}{n}
\sum_{v=1}^n \sum_{u=1}^n a_u b_{\pi^{(v)}(u)} & \geq \frac{1}{n}
\sum_{v=1}^n \sum_{u=1}^n a_u b_u \\
& = \sum_{u=1}^n a_u b_u,
\end{align*}
where $\pi^{(v)}(u) := (u + v) \mod n$ and we have used the
rearrangement inequality for each of these permutations.  \qed \\

\noindent Equipped with these two lemmas, we are now ready to prove
Theorem~\ref{thm:ubeff}.

%%%%%%%%%%%%%%%%%%%%%%%%%%%%%%%%%%%%%%%%%%%%%%%%%%%%%%%%%%%%%%%%%%%%%%%%%%%%

\subsection{Proof of Theorem~\ref{thm:ubeff}}

Without loss of generality, reindexing as necessary, we may assume
that the true permutation $\pi^*$ is the identity $\id$, thereby
ensuring that $M^* = \Mns(\id, \lambda^*)$.  We begin by applying the
triangle inequality to upper bound the error as a sum of two terms:
\begin{align*}
\frac{1}{2} \| \Mhatasp - M^* \|_F^2 & \leq \underbrace{\| \Mhatasp -
  \pihatasp(M^*) \|_F^2}_{\text{estimation error}} + \underbrace{\|
  \pihatasp(M^*) - M^* \|_F^2}_{\text{ approximation error}}.
\end{align*}
Applying Lemma~\ref{lem:scores} yields bound on the approximation
error. In particular, we have
\begin{align*}
\EE \left[ \| \pihatasp(M^*) - M^* \|_F^2 \right] \leq c n \sum_{v \in
  V} \frac{1}{\sqrt{d_v}}.
\end{align*} 

We now turn to the estimation error term, which evaluates to $n^2
(\lambdahat - \lambda^*)^2$, with $\lambdahat$ representing the MLE of
$\lambda^*$ conditional on $\pihat$ being the correct permutation. For
each random set of edges $E$ (we now let $E$ be random in order to
lighten notation) and permutation $\pi$, define the set
\begin{align*}
I_\pi (E) = \{(i,j) \in E \mid i < j, \pi(i) > \pi(j) \},
\end{align*}
corresponding to the set of inversions that are also observed on the
edge set $E$. We require that each ordered pair $(i,j)\in E$ obeys $i
< j$. Therefore, the MLE takes the form
\begin{align*}
1/2 + \lambdahat & = \frac{1}{|E|} \left( \sum_{(i, j) \in E\setminus
  I_{\pihatasp} (E)} Y_{ij} + \sum_{(i,j) \in I_{\pihatasp} (E)} (1 -
Y_{ij}) \right) \\
& = \frac{1}{|E|} \left( \sum_{(i, j) \in E} Y_{ij} + \sum_{(i,j) \in
  I_{\pihatasp} (E)} (1 - 2 Y_{ij}) \right) \\
& = 1/2 + \lambda^* + \frac{1}{|E|} \left( \sum_{(i,j) \in E} W_{ij}
\right) + \frac{1}{|E|} \left( \sum_{(i,j) \in I_{\pihatasp} (E)}
-2\lambda^* - 2 W_{ij} \right),
\end{align*}
where we have written $Y_{ij} = M^*_{ij} + W_{ij}$. Consequently, the
error obeys
\begin{align*}
(\lambdahat - \lambda^*)^2 &\leq \frac{3}{|E|^2} \left( \sum_{(i,j)
    \in E} W_{ij} \right)^2 + \frac{12}{|E|^2} (\lambda^*)^2
  |I_{\pihatasp}(E)|^2 + \frac{12}{|E|^2} \left( \sum_{(i,j)\in
    I_{\pihatasp} (E)} W_{ij} \right)^2 \\
& \stackrel{\1}{\leq} \underbrace{\frac{3}{|E|^2} \left( \sum_{(i,j)
      \in E} W_{ij} \right)^2}_{T_1} + \underbrace{\frac{12}{|E|}
    (\lambda^*)^2 |I_{\pihatasp}(E)|}_{T_2} +
  \underbrace{\frac{12}{|E|^2} \left( \sum_{(i,j)\in I_{\pihatasp}
      (E)} W_{ij} \right)^2}_{T_3},
\end{align*}
where step $\1$ follows since $|I_{\pihatasp}(E)| \leq |E|$
pointwise. We now bound each of the terms $T_1$, $T_2$ and $T_3$
separately. First, by standard sub-exponential tail bounds, and noting
that $W_{ij} \in [-1,1]$, we have
\begin{align*}
\EE [T_1] \leq \frac{3}{|E|}, \quad \text{and} \quad \Pr \left\{ T_1 \geq
\frac{6}{|E|} \right\} \leq e^{-|E|}.
\end{align*}

We also have
\begin{multline*}
\frac{|E|}{12 (\lambda^*)^2} \EE \left[ T_2 \right] = \EE \left[
  |I_{\pihatasp}(E)| \right] \\
 = \sum_{i < j}\sum_{(u,v) \in E} \Pr[\sigma(i)=u,\sigma(j)=v]
\Pr[\pihatasp(i) > \pihatasp(j) |\sigma(i)=u,\sigma(j)=v] \\
 = \sum_{(u,v) \in E} \sum_{i < j} \frac{1}{n(n-1)} \Pr[\pihatasp(i)
  > \pihatasp(j)|\sigma(i)=u,\sigma(j)=v].
\end{multline*}
We now require the following lemma, which is proved at the end of this
section.
\begin{lemma} \label{lem:ktprob}
For any pair of vertices $u \neq v$, we have
\begin{align}
\sum_{i < j} \frac{1}{n(n-1)} \Pr[\pihatasp(i) >
  \pihatasp(j)|\sigma(i)=u,\sigma(j)=v] \leq \frac{c}{\lambda^*}
\left( \frac{1}{\sqrt{d_u}} +
\frac{1}{\sqrt{d_v}}\right). \label{eq:lem1}
\end{align}
\end{lemma}

Using Lemma \ref{lem:ktprob} in conjunction with our previous bounds
yields
\begin{align}
  \label{eq:t2ub}
\EE[T_2] \leq c \frac{\lambda^*}{|E|} \sum_{(u,v) \in E}
\left(\frac{1}{\sqrt{d_u}} + \frac{1}{\sqrt{d_v}} \right) = c
\lambda^* \frac{\sum_{u \in V} \sqrt{d_u}}{\sum_{u \in V} d_u},
\end{align}
where the equality follows since each term $\frac{1}{\sqrt{d_u}}$
appears $d_u$ times in the sum over all edges, and $2|E| = \sum_{u \in
  V} d_u$.  Let $\{d_{(u)}\}_{u=1}^n$ represent the sequence of vertex
degrees sorted in ascending order.  An application of
Lemma~\ref{lem:rearrange} with $a_u = d_{(u)}$ and $b_u =
\frac{1}{\sqrt{d_{(u)}}}$ for $u \in [n]$ yields
\begin{align*}
\sum_{u \in V} \sqrt{d_u} \leq \frac{1}{n}\Big(\sum_{u \in V}
d_u\Big)\Big(\sum_{u \in V} \frac{1}{\sqrt{d_u}}\Big).
\end{align*}
Together with equation~\eqref{eq:t2ub}, we find that
\begin{align*}
\E[T_2] \leq \frac{c \lambda^*}{n} \sum_{u \in V}
\frac{1}{\sqrt{d_u}}.
\end{align*}

In order to complete the proof, it remains to bound $\EE [T_3]$. Note
that this step is non-trivial, since the noise terms $W_{ij}$ for
$(i,j) \in I_{\pihatasp} (E)$ depend on and are coupled through the
data-dependent quantity $\pihatasp$. In order to circumvent this
tricky dependency, consider some \emph{fixed} permutation $\pi$, and
let $T^\pi_3 = \left( \sum_{(i,j)\in I_{\pi} (E)} W_{ij} \right)^2$.
Note that $T^\pi_3$ has two sources of randomness: randomness in the
edge set $E$ and randomness in observations.  Since the observations
$\{W_{ij}\}$ are independent and bounded and $|I_{\pi}(E)| \leq |E|$,
the term
\begin{align*}
\sum_{(i,j)\in I_{\pi} (E)} W_{ij}
\end{align*}
is sub-Gaussian with parameter at most $\sqrt{|E|}$.  We then have the
uniform sub-exponential tail bound
\begin{align}
\Pr \{ T_3^\pi \geq |E| + \delta  \} \leq e^{-c \delta}.
\end{align}

Notice that for any $\alpha \in \mathbb{R}$, the inequality $T_3 \geq
\alpha$ implies that the inequality \mbox{$\frac{12}{|E|^2} T^\pi_3
  \geq \alpha$} holds for some \textit{fixed} permutation $\pi$.
Taking a union bound over all $n! \leq e^{n\log n}$ fixed
permutations, and setting $\delta = c n \log n$ for a constant $c > 1$
yields
\begin{align}
\Pr \left\{ T_3 \geq \frac{12}{|E|} + c \frac{n \log n}{|E|^2}
\right\} \leq \exp \left\{n \log n - c n \log n \right\} \leq \exp
\left\{- c' n \log n \right \}.
\end{align}
Noticing that $T_3 \leq 1$, we obtain
\begin{align*}
\EE [T_3] &\leq \Pr \left\{ T_3 \geq \frac{12}{|E|} + c \frac{n \log
  n}{|E|^2} \right\} + \left(1 - \Pr \left\{ T_3 \geq \frac{12}{|E|} +
c \frac{n \log n}{|E|^2} \right\}\right) \Big(\frac{12}{|E|} + c
\frac{n \log n}{|E|^2}\Big) \\
& \leq \exp \left\{- c' n \log n \right \} + \frac{12}{|E|} + c
\frac{n \log n}{|E|^2} \\
& \leq c' \Big(\frac{1}{|E|} + \frac{n \log n}{|E|^2}\Big).
\end{align*}
Combining the pieces proves the claimed bound on the expectation.
\qed \\

\noindent The only remaining detail is to prove
Lemma~\ref{lem:ktprob}.

%%%%%%%%%%%%%%%%%%%%%%%%%%%%%%%%%%%%%%%%%%%%%%%%%%%%%%%%%%%%%%%%%%%%%%%%%%%%%%%%%

\subsubsection{Proof of Lemma~\ref{lem:ktprob}}

We fix $i,j \in [n]$ with $i<j$ and condition on the event that
$\sigma(i) = u$ and $\sigma(j) = v$ throughout the proof.  First, note
that the bound stated is trivially true if one of the vertices $u$ or
$v$ has degree $1$, by adjusting the constant appropriately. Hence, we
assume for the rest of the proof that $d_u, d_v \geq 2$.  Define the
quantity
\begin{align}
\widetilde{\Delta}_{ji} = 2 \lambda^* \frac{j - i - 1}{n-2}.
\end{align}
We divide the rest of our analysis into two cases.

\paragraph{Case 1, $(u, v) \notin E(G)$:}
When the vertices $u$ and $v$ are not connected, we have
\begin{align*}
\bar{\tau}_j &:= \EE [\widehat{\tau}_j] = \frac 12 + \lambda^*
\Big(\frac{n-j}{n-2} - \frac{j-2}{n-2}\Big) \text{ and}
\\ \bar{\tau}_i &:= \EE [\widehat{\tau}_i] = \frac 12 + \lambda^*
\Big(\frac{n-i-1}{n-2} - \frac{i-1}{n-2}\Big) ,
\end{align*}
and it can be verified that $\bar{\tau}_i - \bar{\tau}_j =
\Deltatilde_{ji}$.  Consequently, we have
\begin{align}
&\Pr \left\{\pihatasp(j) < \pihatasp(i) \mid \sigma(i) = u, \sigma(j)
  = v \right\} \nonumber \\
& \qquad \qquad \qquad= \Pr \left\{\widehat{\tau}_j > \widehat{\tau}_i
  \mid \sigma(i) = u, \sigma(j) = v \right\} \notag \\
& \qquad \qquad \qquad\leq \Pr \left\{|\widehat{\tau}_j -
  \bar{\tau}_j| > \frac{\sqrt{d_u}}{\sqrt{d_v} + \sqrt{d_u}}
  \widetilde{\Delta}_{ji} \;\mid\; \sigma(i) = u, \sigma(j) = v
  \right\} \notag \\
& \qquad \qquad \qquad \qquad+ \Pr \left\{|\widehat{\tau}_i -
  \bar{\tau}_i| > \frac{\sqrt{d_v}}{\sqrt{d_v} + \sqrt{d_u}}
  \Deltatilde_{ji} \;\mid\; \sigma(i) = u, \sigma(j) = v \right\}
  \notag \\
  & \qquad \qquad \qquad\leq 4 \exp\left\{ - c \frac{d_u
    d_v}{(\sqrt{d_u} + \sqrt{d_v})^2} \widetilde{\Delta}_{ji}^2
  \right\}, \label{eq:case1}
\end{align}
where the last step follows from the Hoeffding bound for sampling
without replacement in conjunction with the standard Hoeffding bound
for bounded independent noise, by an argument similar to that of
equation~\eqref{eq:hoeffding}.

%%%%%%%%%%%%%%%%%%%%%%%%%%%%%%%%%%%%%%%%%%%%%%%%%%%%%%%%%%%%%%%%%%%%%%%%%%%%%%

\paragraph{Case 2, $(u, v) \in E(G)$:}

When the vertices $u$ and $v$ are connected, we have
\begin{align*}
\bar{\tau}_j &:= \EE [\widehat{\tau}_j] = \frac 12 + \frac{d_v -
  1}{d_v} \lambda^* \Big( \frac{n-j}{n-2} - \frac{j-2}{n-2} \Big) -
\frac{1}{d_v} \lambda^* \text{ and}\\ \bar{\tau}_i &:= \EE
     [\widehat{\tau}_i] = \frac 12 + \frac{d_u - 1}{d_u} \lambda^*
     \Big(\frac{n-i-1}{n-2} - \frac{i-1}{n-2}\Big) + \frac{1}{d_u}
     \lambda^*,
\end{align*}
and it can be verified that $\bar{\tau}_i - \bar{\tau}_j \geq
\widetilde{\Delta}_{ji}$.

Now, however, we must apply the Hoeffding bound for sampling without
replacement to $d_u - 1$ and $d_v - 1$ random variables, respectively.
Recalling that $d_u, d_v \geq 2$, we have
\begin{align}
& \Pr \left\{\pihatasp(j) < \pihatasp(i) \mid \sigma(i) = u, \sigma(j)
  = v \right\} \notag \\ &\qquad \qquad \qquad= \Pr
  \left\{\widehat{\tau}_j > \widehat{\tau}_i \mid \sigma(i) = u,
  \sigma(j) = v \right\} \notag \\ &\qquad \qquad \qquad\leq \Pr
  \left\{|\widehat{\tau}_j - \bar{\tau}_j| >
  \frac{\sqrt{d_u}}{\sqrt{d_v} + \sqrt{d_u}} \widetilde{\Delta}_{ji}
  \;\mid\; \sigma(i) = u, \sigma(j) = v \right\} \notag \\ &\qquad
  \qquad \qquad \qquad + \Pr \left\{|\widehat{\tau}_i - \bar{\tau}_i|
  > \frac{\sqrt{d_v}}{\sqrt{d_v} + \sqrt{d_u}} \Deltatilde_{ji}
  \;\mid\; \sigma(i) = u, \sigma(j) = v \right\} \notag \\ &\qquad
  \qquad \qquad \leq 4 \exp\left\{ - c \frac{(d_u - 1) (d_v -
    1)}{(\sqrt{d_u -1} + \sqrt{d_v -1})^2} \widetilde{\Delta}_{ji}^2
  \right\} \notag \\ &\qquad \qquad \qquad\leq 4 \exp\left\{ - c'
  \frac{d_u d_v}{(\sqrt{d_u} + \sqrt{d_v})^2}
  \widetilde{\Delta}_{ji}^2 \right\}. \label{eq:case2}
\end{align}

We use the shorthand $L_{uv}$ to denote the LHS of
equation~\eqref{eq:lem1}. Having established the
bounds~\eqref{eq:case1} and~\eqref{eq:case2}, we now combine them to
derive that
\begin{align}
L_{uv} & \leq \frac{1}{n(n-1)} \sum_{j =2}^{n} \sum_{i < j} 4
\exp\left\{ - c \frac{d_u d_v}{(\sqrt{d_u} + \sqrt{d_v})^2} (j - i
-1)^2 \frac{(\lambda^*)^2}{(n-2)^2}\right\} \notag \\
& \leq \frac{4}{n(n-1)} (n-1) \sum_{m = 1}^n \exp\left\{ - \frac{d_u
  d_v}{(\sqrt{d_u} + \sqrt{d_v})^2} m^2
\frac{(\lambda^*)^2}{(n-2)^2}\right\}, \notag
\end{align}
where we have used $m = j - i$, and noted that there are at most $n -
1$ repetitions of each distinct value of $j - i$ in the sum over $j >
i$.

Defining $\psi(q) = \sum_{m=1}^\infty q^{m^2}$, we recall the
following theta function identity\footnote{For the rest of this
  subsection, $\pi$ denotes the universal constant.} for $ab = \pi$
(see, for instance, equation (2.3) in Yi~\cite{yi04theta}):
\begin{align*}
\sqrt{a} \left(1 + 2\psi(e^{-a^2})\right) = \sqrt{b} \left( 1 + 2
\psi(e^{-b^2})\right).
\end{align*}
Using the identity by setting $a^2 = c \frac{d_u d_v}{(\sqrt{d_u} +
  \sqrt{d_v})^2} \frac{(\lambda^*)^2}{n^2}$ yields
\begin{align}
L_{uv} &\leq \frac{c}{n} \frac{n}{\lambda^*} \frac{\sqrt{d_u} +
  \sqrt{d_v}}{\sqrt{d_u d_v}} \left( 1 + 2\sum_{m = 1}^\infty
\exp\left\{ - \pi^2 \frac{(\sqrt{d_u} + \sqrt{d_v})^2}{d_u d_v} m^2
\frac{n^2}{(\lambda^*)^2} \right\} \right)\notag \\
& \leq \frac{c}{\lambda^*} \frac{\sqrt{d_u} + \sqrt{d_v}}{\sqrt{d_u
    d_v}} \left( 1 + 2 \sum_{m = 1}^\infty \exp\left\{ - \pi^2
\frac{(\sqrt{d_u} + \sqrt{d_v})^2}{d_u d_v} m
\frac{n^2}{(\lambda^*)^2} \right\} \right)\notag \\
& \leq \frac{c}{\lambda^*} \frac{\sqrt{d_u} + \sqrt{d_v}}{\sqrt{d_u
    d_v}} \left(1 + \sum_{m = 1}^\infty \exp\left\{ - 16 \pi^2 n m
\right\} \right),
\end{align}
where in the last step, we have used the fact that $\lambda^* \leq
1/2$, and that $\frac{(\sqrt{d_u} + \sqrt{d_v})^2}{d_u d_v} \geq
4/n$. Bounding the geometric sum by a universal constant yields the
required result.

%%%%%%%%%%%%%%%%%%%%%%%%%%%%%%%%%%%%%%%%%%%%%%%%%%%%%%%%%%%%%%%%%%%%%%%%%%%%%%%%

\subsection{Proof of Theorem~\ref{thm:lb1}}
We prove the two parts of the theorem separately.

\subsubsection{Proof of part (a)}

The proof of part (a) is based on the following lemmas.
\begin{lemma}
  \label{lem:lb1}
Consider a matrix of the form $M^* = \Mns(\pi^*, 1/4)$ where the
permutation $\pi^*$ is chosen uniformly at random. For any graph $G = K_1 \cup K_2 \cup \ldots$
composed of multiple disjoint cliques with the number of vertices bounded as $C \leq |K_i| \leq n/5$ for all $i$, and for any estimators $(\Mhat,
\pihat)$ that are measurable functions of the observations on $G$, we
have
\begin{align}
\EE \left[ \frac{1}{n^2} \|\Mhat - M^* \|_F^2 \right] \geq
\frac{c_2}{n} \sum_{v \in V} \frac{1}{\sqrt{d_v}}, \quad \text{ and }
\quad \EE \left[ \KT(\pi^*, \pihat) \right] \geq c_2 n \sum_{v \in V}
\frac{1}{\sqrt{d_v}}.
\end{align}
\end{lemma}

\begin{lemma}
  \label{lem:clique}
Given any graph $G$ with degree sequence $\{d_v\}_{v \in V}$, there
exists a graph $G'$ consisting of multiple disjoint cliques with
degree sequence $\{d_v'\}_{v \in V}$ such that
\begin{align}
|E| \asymp |E'| \quad \text{ and } \quad \sum_{v \in V}
\frac{1}{\sqrt{d_v}} \asymp \sum_{v \in V} \frac{1}{\sqrt{d_v'}}.
\end{align}
\end{lemma}
\noindent Part (a) follows by combining these two lemmas, so that it
suffices to prove each of the lemmas individually.

%%%%%%%%%%%%%%%%%%%%%%%%%%%%%%%%%%%%%%%%%%%%%%%%%%%%%%%%%%%%%%%%%%%%%%%%%%%%%%%

\paragraph{Proof of Lemma~\ref{lem:lb1}:}

Our result is structural, and proved for permutation recovery.  The
bound for matrix recovery follows as a corollary.  Assume we are given
a graph on $n$ vertices consisting of $k$ disjoint cliques of sizes
$n_1, \dots, n_k$.  Let $N_0=0$ and $N_j = \sum_{i=1}^{j} n_i$ for $j
\in [k]$.  Without loss of generality, we let the $j$-th clique
consist of the set of vertices $V_j$ indexed by $\{N_{j-1} + 1,
\ldots, N_j\}$. By assumption, each $n_j$ is upper bounded by $n/5$ and lower bounded by a universal constant. %, as the extension to the general case is straightforward.

Note that any estimator can only use the observations to construct the
correct partial order within each clique, but not across cliques.  We
denote the induced partial order of a permutation $\pi$ on the clique
$V_j$ by the permutation $\pi_j: [n_j] \to [n_j]$\footnote{As an
  example, the identity permutation $\pi = \id$ would yield $\pi_j =
  \id$ on $[n_j]$ for all $j \in [k]$.}.  We will demonstrate that
there exists a coupling of two marginally uniform random permutations
$(\pi^*, \pi^\#)$ such that
\begin{align*} 
\EE [\KT(\pi^*, \pi^\#)] \geq c n \sum_{j=1}^k \sqrt{n_j} = cn \sum_{v
  \in V} \frac{1}{\sqrt{d_v}} ,
\end{align*} 
and the partial order of $\pi^*$ agrees with that of $\pi^\#$ on each
clique, that is, $\pi^*_j = \pi^\#_j \text{ for all } j \in [k]$.
Another way of stating this is that for every clique $V_j$ and every
two vertices $i_1,i_2 \in V_j$, we need that $\pi^\#(i_1) <
\pi^\#(i_2)$ if and only if $\pi^*(i_1) < \pi^*(i_2)$.

Let $\EE[ \cdot \mid \pi^*]$ denote the expectation over the
observations conditional on $\pi^*$. Given a pair of permutations $(\pi^*, \pi^\#)$
satisfying the above assumption, we view them as two hypotheses of the
latent permutation. Then for any estimator $\pihat$, the
Neyman-Pearson lemma~\cite{neyman1966joint} guarantees that
\begin{align*} 
\EE[ \KT(\pihat, \pi^*) \mid \pi^*] + \EE[ \KT(\pihat, \pi^\#) \mid
  \pi^\# ] \ge \KT(\pi^\#, \pi^*)
\end{align*}
for each instance of $(\pi^*,\pi^\#)$, because the observations are
identical for $\pi^*$ and $\pi^\#$. Taking expectation over
$(\pi^*,\pi^\#)$, we obtain that
\begin{align*} 2\, \EE [\KT(\pihat, \pi^*) ] \geq \EE[ \KT(\pi^*, \pi^\#) ] \ge c n  \sum_{v \in V} \frac{1}{\sqrt{d_v}} \end{align*}
since both $\pi^*$ and $\pi^\#$ are marginally uniform.

To finish the proof, it remains to construct the required coupling
$(\pi^*, \pi^\#)$.  The construction is done as follows.  First,
permutations $\pi^*$ and $\tilde \pi$ are generated uniformly at
random and independently.  Second, we sort the permutation $\tilde
\pi$ on each clique according to $\pi^*$, and denote the resulting
permutation by $\pi^\#$. Then the permutations $\pi^*$ and $\pi^\#$
are marginally uniform and have common induced partial orders on the
cliques, which we denote by $\{\pi^*_j: j \in [k]\}$.

With some extra notation, we can define the
sorting step more formally for the interested reader. For a set of
partial orders on the cliques $\{\pi_j: j \in [k]\}$, we define a
special permutation that effectively orders vertices within each
clique $V_j$ according to its corresponding partial order $\pi_j$, but
does not permute any vertices across cliques.  We denote this special
permutation by $\pi_{\sf par}(\{\pi_j: j \in [k]\})$.  For every
clique $V_j$, we consider the permutation $\pi_{{\sf sort}, j} :=
\pi^*_j \circ (\tilde \pi_j)^{-1}$.  Now, we can formally define the
sorting step to generate $\pi^\#$ by
\begin{align*}
\pi^\# \defn \pi_{\sf par}(\{\pi_{{\sf sort}, j}: j \in [k]\}) \circ
\tilde \pi .
\end{align*}

Next, we need to evaluate the expected Kendall's tau distance between
these coupled permutations. By the tower property, we have
\begin{align*}
\EE[\KT(\pi^*, \pi^\#)] &= \EE \big[ \EE \big[\KT(\pi^*,\pi^\#) \mid
    \{\pi^*_j: j \in [k]\} \big] \big] .
\end{align*}
The inner expectation can be simplified as follows. Pre-composing
permutations $\pi^*$ and $\pi^\#$ with any permutation does not change
the Kendall's tau distance between them, so we have
\begin{align*}
\EE \big[\KT(\pi^*,\pi^\#) \mid \{\pi^*_j : j \in [k]\} \big] =
\EE[\KT(\pi,\pi')]
\end{align*}
where the permutations $\pi$ and $\pi'$ are drawn independently and
uniformly at random from the set of permutations that are increasing
on every clique.  That is, for every clique $V_j$ and every two
vertices $i_1,i_2 \in V_j$, we have\footnote{To understand why $\pi$
  and $\pi'$ can be chosen independently, note that the only
  dependency between the original permutations $\pi^*$ and $\pi^\#$ is
  through the common induced partial orders $\{\pi^*_j: j \in
  [k]\}$. By conditioning and pre-composing, we are able to remove
  that dependency.} $\pi(i_1) < \pi(i_2)$ and $\pi'(i_1) < \pi'(i_2)$.

We now turn to computing the quantity $\EE[\KT(\pi,\pi')]$.  It is
well-known~\cite{diaconis1977spearman} that $2 \, \KT(\pi,\pi') \ge
\|\pi-\pi'\|_1$.  This fact together with Jensen's inequality implies
that
\begin{align} 
2 \, \EE[\KT(\pi,\pi')] & \geq \sum_{i=1}^n \EE\big[ |\pi(i) -
  \pi'(i)| \big] \nonumber\\
& \ge \sum_{i=1}^n \EE \Big[ \Big| \E \big[ \pi(i) - \pi'(i) \mid \pi
    \big] \Big| \Big] \nonumber\\
& = \sum_{i=1}^n \EE \Big[ \Big| \pi(i) - \E[\pi'(i)] \Big| \Big]
\nonumber\\
&= \EE \big[ \big\| \pi-\E[\pi] \big\|_1 \big]
. \label{eq:spearman_jensen}
\end{align}
It therefore suffices to lower bound the quantity $\EE[\vecnorm{\pi -
    \E[\pi]}{1}]$.

Fix any $i \in [n]$. Then $i$ is $\ell$-th smallest index in the
$j$-th clique for some $j \in [k]$ and $\ell \in [n_j]$, or
succinctly, $i = N_{j-1} + \ell$.  If we view $\pi^{-1}$ as random
draws from the $n$ items, then $\pi(i)$ is equal to the the number of
draws needed to get the $\ell$-th smallest element of $V_j$.  Denoting
$\E[\pi(i)]$ by $\mu$, we have
\begin{align*} 
\mu = \ell + \EE\Big[ \sum_{r: \sigma(r) \notin V_j} \mathbf{1}\big\{r
  \text{ is drawn before } i \big\} \Big] = \ell + (n-n_j)
\frac{\ell}{n_j+1} = \ell \frac{n+1}{n_j+1},
\end{align*}
since the probability that an item not in $V_j$ is drawn before the
$\ell$-th smallest element of $V_j$ is $\ell/(n_j+1)$.  Furthermore,
$\pi(i) = s$ if and only if $\ell-1$ elements of $V_j$ are selected in
the first $s-1$ draws and the $s$-th draw is from $V_j$, so
\begin{align} \label{eq:pmf}
  \Pr \{\pi(i) = s\} = \binom{n_j}{\ell-1} \binom{n-n_j}{s-\ell}
  \binom{n}{s-1}^{-1}\, \frac{n_j-\ell+1}{n-s+1}.
\end{align}
We claim that for all $\lceil 2n_j/5 \rceil \le \ell \le \lfloor 3 n_j
/5 \rfloor$ and $|s-\mu|\le n/\sqrt{n_j}$, it holds that
\begin{align} \label{eq:prob-bd}
\Pr \{\pi(i) = s\} \le c \sqrt{n_j}/n
\end{align}
where $c$ is a universal positive constant.

If the claim holds, then for any $0 \le m \le n/\sqrt{n_j}$, we have
\begin{align*} 
\EE\big[ |\pi(i) - \mu| \big] & \ge m \Pr \big\{ |\pi(i)-\mu| \ge m
\big\} \ge m \big[ 1- c (2m+1) \sqrt{n_j}/n \big]
\end{align*}
by Markov's inequality. Choosing $m = \frac{n}{6 c \sqrt{n_j}}$ yields
\begin{align*} 
\EE\big[ |\pi(i) - \mu| \big] & \ge c_2 n/\sqrt{n_j}
\end{align*}
for some positive constant $c_2$. Summing over $\ell$ in the given
range, together with inequality~\eqref{eq:spearman_jensen}, completes
the proof.

\paragraph{Proof of claim~\eqref{eq:prob-bd}:}

For $\ell \in [n_j]$ and $\ell \le s \le n-n_j+\ell$, define a
bivariate function
\begin{align*}
p(\ell, s) \defn \binom{n_j}{\ell-1} \binom{n-n_j}{s-\ell}
\binom{n}{s-1}^{-1} .
\end{align*}
Note that for any fixed $s$, the function $\ell \mapsto p(\ell,s)$ is
the probability mass function of the hypergeometric distribution that
describes the probability of $\ell-1$ successes in $s-1$ draws without
replacement from a population of size $n$ with $n_j$ successes. Hence,
its maximum is attained at $\ell = \big \lfloor s \frac{n_j+1}{n+2}
\big \rfloor$. Now we consider the index set
\begin{align*} 
\mathcal I = \Big\{(l, s) : \Big\lceil \frac{n_j}3 \Big\rceil \le \ell
\le \Big\lceil \frac{2 n_j}3 \Big\rceil, \Big\lceil \frac{n_j}3
\Big\rceil \le \Big \lfloor s \frac{n_j+1}{n+2} \Big \rfloor \le
\Big\lceil \frac{2 n_j}3 \Big\rceil \Big\} \subset \Big[\frac{n_j}3,
  \frac{2n_j}{3} \Big] \times \Big[\frac{n}5, \frac{4n}{5} \Big] .
\end{align*}
In particular, the range of interest $\lceil 2 n_j/5 \rceil \le \ell
\le \lfloor 3 n_j /5 \rfloor$ and $|s-\mu|\le n/\sqrt{n_j}$, is
contained within the set $\mathcal I$, since $\mu = \ell
\frac{n+1}{n_j+1}$. Moreover, inequality \eqref{eq:pmf} ensures that
$\Pr\{\pi(i) = s\} \le p(\ell,s) \frac{c_1 n_j}n$ for $(\ell,s) \in
\mathcal I$. Thus, in order to complete the proof, it suffices to
prove that $p(\ell,s) \le c /\sqrt{n_j}$ for $(\ell, s) \in \mathcal
I$, and it suffices to consider $(\ell,s)$ such that $\ell = \big
\lfloor s \frac{n_j+1}{n+2} \big \rfloor$ since each function $\ell
\mapsto p(\ell,s)$ attains its maximum at such a pair $(\ell,s)$.

Toward this end, we use Stirling's approximation \cite{de1756doctrine}
to obtain
\begin{align} 
p(\ell,s) &\le c_2 \frac{\sqrt{n_j (n-n_j)
    (s-1) (n-s+1)}}{\sqrt{(\ell-1) (n_j - \ell+1) (s-\ell)
    (n-n_j-s+\ell) n}} \label{eq:ratio} \\
& \quad \cdot \frac{n_j^{n_j} (n-n_j)^{n-n_j} (s-1)^{s-1}
  (n-s+1)^{n-s+1}}{(\ell-1)^{\ell-1} (n_j - \ell+1)^{n_j-\ell+1}
  (s-\ell)^{s-\ell} (n-n_j-s+\ell)^{n-n_j-s+\ell}
  n^n} . \label{eq:stir-exp}
\end{align}
Since the factor in line \eqref{eq:ratio} scales as $1/\sqrt{n_j}$ for $(\ell,s) \in \mathcal I$, it remains to bound the factor in line \eqref{eq:stir-exp} by a universal constant. This follows from lengthy yet standard approximations which we briefly describe here. Assume that $s \frac{n_j+1}{n+2}$ is an integer for simplicity, so that $\ell$ is equal to this quantity and we have $s = \ell \frac{n+2}{n_j+1}$; the extension to the general case is easy.
We first group together
\begin{align*} 
\Big[\frac{n_j (s-1)}{(\ell-1) n}\Big]^{\ell-1} &= \Big[ \frac{n_j (n \ell + 2 \ell - n_j - 1)/(n_j+1)}{(\ell-1) n} \Big]^{\ell-1} \\
%&= \Big[1+ \frac{n_j (n \ell + 2 \ell - n_j - 1)/(n_j+1)/n-\ell+1}{\ell-1} \Big]^{\ell-1} \\
&= \Big[1+ \frac{1+(2\ell n_j - n_j^2 - n_j - \ell n)/(n_j n+ n)}{\ell-1} \Big]^{\ell-1} ,
\end{align*}
which is bounded by a constant for $(\ell,s) \in \mathcal I$ considering that $\lim_{m \to \infty} (1+\frac{a}{m})^m = e^a$.
%because of the assumptions $n_j/3 \le \ell \le 2 n_j/3$ and $n_j \le n/3$. 
Then, we
group together the terms
\begin{align*}
\Big[ \frac{n_j (n-s+1)}{(n_j-\ell+1) n} \Big]^{n_j-\ell+1}, \qquad
\Big[ \frac{(n-n_j) (s-1)}{(s-\ell) n} \Big]^{s-\ell} \quad
\text{and} \quad \Big[ \frac{(n-n_j) (n-s+1)}{(n-n_j-s+\ell) n}
  \Big]^{n-n_j-s+\ell}
\end{align*}
respectively, and a similar argument yields that each term is
bounded by a constant.  \qed

%%%%%%%%%%%%%%%%%%%%%%%%%%%%%%%%%%%%%%%%%%%%%%%%%%%%%%%%%%%%%%%%%%%%%%%

\paragraph{Proof of Lemma~\ref{lem:clique}:}

Fix a graph $G$ with degree sequence $\{d_v\}_{v \in V}$, and
introduce the shorthand $S = \sum_{v \in V} 1/\sqrt{d_v}$. For some
parameter $k$ to be chosen, define the graph $G'$ on the same vertex
set to be the disjoint union of one clique of size $c_1
\lfloor\sqrt{|E|}\rfloor$, $c_2 k$ cliques of size $\lfloor n/k
\rfloor$ and $c_3 S$ cliques of size $2$, where $c_1, c_2$ and $c_3$
are constants to be determined such that the sizes of each clique are
integers. The number of vertices remains the same, so that
\begin{align}
  \label{eq:vert-same} 
n = c_1 \lfloor \sqrt{|E|}\rfloor + c_2 k \lfloor n/k \rfloor + 2
  c_3 S.
\end{align}
The number of edges of $G'$ is
\begin{align*}
  |E'| = \binom{c_1\lfloor\sqrt{|E|}\rfloor}{2} + c_2 k \binom{\lfloor
    n/k \rfloor}{2} + c_3 S \; \asymp \; |E| +
  \frac{n^2}{k},
\end{align*}
where the last approximation holds because $S\le n \le 2 |E|$.
Moreover, let
\begin{align*}
  S' = \sum_{v \in V} \frac{1}{\sqrt{d_v'}} =
  \frac{c_1\lfloor\sqrt{|E|}\rfloor}{\sqrt{c_1\lfloor\sqrt{|E|}\rfloor-1}}
  + \frac{c_2 k \lfloor n/k \rfloor}{\sqrt{\lfloor n/k \rfloor -1}} +
  c_3 S \; \asymp \; \sqrt{nk} + S,
\end{align*}
where the last approximation holds since $|E|^{1/4} \le \sqrt n \le
S$.

In order to guarantee that $|E'| \asymp |E|$ and $S' \asymp S$, we
need to choose an integer $k$ so that $n^2/k \le c|E|$ and $\sqrt{nk}
\le cS$, or equivalently
\begin{align*}
  \frac{n^2}{c|E|} \le k \le c^2 \frac{S^2}{n}.
\end{align*}
Such an integer $k$ exists if $|E|S^2 \ge n^3$. Indeed, applying
Lemma~\ref{lem:rearrange} twice (with $a_u = d_{(u)}$ and $b_u
=1/\sqrt{d_{(u)}}$ the first time and $a_u = \sqrt{d_{(u)}}$ and $b_u
=1/\sqrt{d_{(u)}}$ the second time, where $\{d_{(u)}\}_{u=1}^n$ is the
degree sequence in ascending order), we obtain that
\begin{align*}
  |E| S^2 = \Big(\sum_{v \in V} d_v \Big) \Big(\sum_{v \in V}
  \frac{1}{\sqrt{d_v}}\Big)^2 \ge n \Big(\sum_{v \in V} \sqrt{d_v}
  \Big) \Big(\sum_{v \in V} \frac{1}{\sqrt{d_v}}\Big) \ge n^3.
\end{align*}
With $k$ selected, it is easy to choose $c_1, c_2$ and $c_3$ so that
inequality~\eqref{eq:vert-same} holds, since each of $\sqrt{|E|}$, $k
\lfloor n/k \rfloor$ and $S$ is no larger than $n$. The issue of
integrality can be taken care of by constant-order adjustment of these
numbers, so the proof is complete.  \qed

%%%%%%%%%%%%%%%%%%%%%%%%%%%%%%%%%%%%%%%%%%%%%%%%%%%%%%%%%%%%%%%%%%%%%%%%%%

\subsubsection{Proof of part (b)}

Given a parameter space $\Theta$, a set $\mathcal{P} = \{ \theta_1,
\theta_2, \ldots, \theta_{|\mathcal{P}|} \}$ is said to be a
$\delta$-packing in the metric $\rho$ if $\rho(\theta_i, \theta_j) >
\delta$ for all $i \neq j$.  The lower bound of part (b) is based on
the following packing lemma for the set of permutations in Kendall's
tau distance. We note that a similar lemma was proved by Barg and
Mazumdar~\cite{barg2010codes}.

\begin{lemma}
\label{lem:ktpack}
For some positive constant $c_1$, there exists an $c_1 n^2$-packing
$\mathcal P$ of the set of permutations in the Kendall's tau distance
such that $\log |\mathcal P| \ge n$.
\end{lemma}

Consider the random observation model with graph $G = (V,E)$, where
$E$ denotes the random edge set of observations.  We denote by
$\mathbb{Q}_{M}$ the law of the random observation noisy sorting model
with underlying matrix $M = \Mns(\pi, \lambda)$. We require the
following lemma.

\begin{lemma}
\label{lem:nskl}
Let $\mathbb{P}_{M,G}$ denote the law of the noisy sorting model with
underlying matrix \mbox{$M \in \Cns(\lambda)$} for $\lambda \in
[0,1/4]$ and comparison graph $G$. Suppose that the entries of two
matrices $M, M' \in \Cns(\lambda)$ differ in $s$ edges of the graph
$G$. Then the KL divergence is bounded as
\begin{align}
  \mathsf{KL}(\mathbb{P}_{M,G}, \mathbb{P}_{M',G}) \le 9 \lambda^2 s.
\end{align}
\end{lemma}

Note that conditional on any instance of $E$, Lemma~\ref{lem:nskl}
guarantees that
\begin{align*}
\mathsf{KL}(\mathbb{P}_{M,G}, \mathbb{P}_{M',G}) \le 9 \lambda^2
\Big|\{(i,j) \in E: i<j, M_{i,j} \ne M'_{i,j} \}\Big| ,
\end{align*}
where $\mathbb{P}_{M,G}$ denotes the model for fixed graph $G$. Hence
taking expectation over the random edge set yields the upper bound
\begin{align*}
\mathsf{KL}(\mathbb{Q}_{M}, \mathbb{Q}_{M'}) \le 9 \lambda^2
\sum_{i<j, \, M_{i,j} \ne M'_{i,j}} \Pr\{(i,j) \in E\} \le 9 \lambda^2
\sum_{i<j} \frac{2 |E|}{n(n-1)} = 9 \lambda^2 |E|,
\end{align*}
valid for any $M,M' \in \Cns(\lambda)$.

Note that $\|M-M'\|_F^2 = 8 \lambda^2 \KT(\pi,\pi')$ for $M =
\Mns(\pi,\lambda)$ and $M' = \Mns(\pi',\lambda)$. Hence Fano's
inequality applied to the packing given by Lemma~\ref{lem:ktpack}
yields that
\begin{align*}
\inf_{\Mhat} \sup_{M^* \in \Cns} \EE \left[ \| \Mhat - M^* \|_F^2
  \right] \geq 8 \lambda^2 c_1 n^2 \left(1 - \frac{9 \lambda^2 |E| +
  \log 2}{n} \right).
\end{align*}
The proof is completed by choosing $\lambda^2 = c_2 n/|E|$ for a
sufficiently small constant $c_2$.  \qed

It remains to prove Lemmas~\ref{lem:ktpack} and~\ref{lem:nskl}.

%%%%%%%%%%%%%%%%%%%%%%%%%%%%%%%%%%%%%%%%%%%%%%%%%%%%%%%%%%%%%%%%%%%%%%%%%%%

\paragraph{Proof of Lemma~\ref{lem:ktpack}:}

The inversion table $b=(b_1, \dots, b_n)$ of a permutation $\pi$ has
entries defined by
\begin{align*} 
b_i = \sum_{j=i+1}^n \mathbf{1} \{\pi(i) > \pi(j)\} \text{ for each }
i \in [n].
\end{align*}
We refer the reader to Mahmoud~\cite{mahmoud00sorting} and references therein for background
on inversion tables.
By definition, we have $b_i \in \{0,1, \dots, n-i\}$ and $\KT(\pi,
\mathsf{id}) = \sum_{i=1}^n b_i$ where $\mathsf{id}$ denotes the
identity permutation. In fact, the set of tables $b$ satisfying $b_i
\in \{0,1, \dots, n-i\}$ is bijective to the set of permutations via
this relation~\cite{mahmoud00sorting}. This bijection aids in counting
permutations with constraints.

Denote by $\mathcal B(\id, r)$ the set of permutations that are within
Kendall's tau distance $r$ of the identity $\id$. We seek an upper
bound on $|\mathcal B(\id, r)|$. Every $\pi \in \mathcal B(\id, r)$
corresponds to an inversion table $b$ such that $\sum_{i=1}^n b_i \le
r$. If $b_i$ is only required to be a nonnegative integer, then the
number of $b$ satisfying $\sum_{i=1}^n b_i \le r$ is bounded by
$\binom{n+r}{n}$. After taking logarithms, this yields a bound
\begin{align*} 
\log |\mathcal B(\id, r)| \le n \log (1+r/n) + n .
\end{align*}

Let $\mathcal P$ be a maximal $c_1 n^2$-packing of the set of
permutations, which is necessarily also a $c_1 n^2$-covering of that
set. Then the family $\{\mathcal B(\pi, c_1 n^2)\}_{\pi \in \mathcal
  P}$ covers all permutations. By the right-invariance of the
Kendall's tau distance under composition, the above bound yields $\log
|\mathcal B(\pi, c_1 n^2)| \le n \log (1+c_1 n) + n$ for each
$\pi$. Since there are $n!$ permutations in total, we conclude that
$\log |\mathcal P| \ge \log(n!) - n \log (1+c_1 n) - n \ge n$ for a
sufficiently small constant $c_1$.  \qed

%%%%%%%%%%%%%%%%%%%%%%%%%%%%%%%%%%%%%%%%%%%%%%%%%%%%%%%%%%%%%%%%%%%%%%%%%%%

\subsubsection{Proof of Lemma~\ref{lem:nskl}}

The KL divergence between Bernoulli observations has the form
\begin{align*}
\mathsf{KL} \big( \BER(1/2 + \lambda), \BER(1/2 - \lambda) \big) &=
\mathsf{KL} \big( \BER(1/2 - \lambda), \BER(1/2 + \lambda) \big) \\
& = (1/2+\lambda) \log \frac{1/2+\lambda}{1/2-\lambda} + (1/2-\lambda)
\log \frac{1/2-\lambda}{1/2+\lambda} \\
& = 2 \lambda \log \frac{1/2+\lambda}{1/2-\lambda} \\
& \leq 9 \lambda^2 \qquad \mbox{for all $\lambda \in [0, 1/4]$,}
\end{align*}
where the last inequality follows by some simple algebra,

Note that the KL divergence between a pair of product distributions is
equal to the sum of the KL divergences between individual pairs.
Since $M$ and $M'$ differ in $s$ entries on the graph $G$ and the
Bernoulli observations are independent for different edges, we see
that $\mathsf{KL}(\mathbb{P}_{M,G}, \mathbb{P}_{M',G}) \le 9 \lambda^2
s.$ \qed

%%%%%%%%%%%%%%%%%%%%%%%%%%%%%%%%%%%%%%%%%%%%%%%%%%%%%%%%%%%%%%%%%%%%%%%%%%%%

\subsection{Proof of Theorem~\ref{thm:sstupper}}

For the purpose of the proof, it is helpful to think of the
observation model in its linearized form. In particular, we have two
random edge sets $E_1$ and $E_2$ and the observation matrices
\begin{align*}
Y_i & \defn  M^* + W_i
\end{align*}
for each $i \in \{1, 2\}$. We also use the shorthand $\Bl(X, C) \defn
\Bl(X, C, [n] \times [n])$, and recall the notation $\|M\|^2_B \defn
\sum_{(i,j) \in B} M_{ij}$.

By the triangle inequality, we have
\begin{align}
\|\Mhatbap - M^* \|_F^2 &\leq 2 \|\Mhatbap - \pihatasp(M^*) \|_F^2 + 2
\| M^* - \pihatasp(M^*) \|_F^2 \notag \\
& \stackrel{\1}{\leq} 2 \|\Mtilde - \pihatasp(M^*) \|_F^2 + 2 \| M^* -
\pihatasp(M^*) \|_F^2 \notag \\
& \leq 4 \|\Mtilde - M^* \|_F^2 + 6 \| M^* - \pihatasp(M^*)
\|_F^2, \label{eq:triangle}
\end{align}
where step $\1$ follows from the non-expansiveness of the projection
operator. We know from Lemma \ref{lem:scores} that the second term in
inequality \eqref{eq:triangle} is bounded in expectation by the
quantity $n S = n \sum_{v \in V} 1 / \sqrt{d_v}$ as desired, so it
remains to bound the first term. Toward that end, again apply triangle
inequality to write
\begin{align}
  \label{eq:sst-est}
\| \Mtilde - M^* \|_F^2 \leq 2 \| \Mtilde - \Bl(M^*, \widehat{b})
\|_F^2 + 2 \| M^* - \Bl(M^*, \widehat{b}) \|_F^2.
\end{align}
We now bound each of these terms separately. Starting with the first,
let us define some notation. For a set $S \subseteq [n] \times [n]$
and a matrix $M \in \real^{n \times n}$, let $\|M\|^2_S = \sum_{(i,j)
  \in S} M^2_{ij}$. We have
\begin{align*}
\| \Mtilde - \Bl(M^*, \widehat{b}) \|_F^2 &= \sum_{B \in
  \mathcal{B}(\bhat)}\| \Mtilde - \Bl(M^*, \widehat{b}) \|_B^2.
\end{align*}

Note that it is sufficient to consider off diagonal blocks in the sum,
since both $\Mtilde$ and $\Bl(M^*, \widehat{b})$ are identically $1/2$
in the diagonal blocks.  Considering each block separately, we now
split the analysis into two cases.

%%%%%%%%%%%%%%%%%%%%%%%%%%%%%%%%%%%%%%%%%%%%%%%%%%%%%%%%%%%%%%%%%%%%%%%%%%%%%

\paragraph{Case 1, $B \cap E_2 = \phi$:}

Because the entries of the error matrix are bounded within $[-1, 1]$, we have
\begin{align*}
\| \Mtilde - \Bl(M^*, \widehat{b}) \|_B^2 \leq |B|.
\end{align*}

\paragraph{Case 2, $B \cap E_2 \neq \phi$:}

Since both $\Mtilde$ and $\Bl(M^*, \widehat{b})$ are constant on each
block, we have
\begin{align}
\| \Mtilde - \Bl(M^*, \widehat{b}) \|_B^2 &= \frac{|B|}{|B \cap E_2|}
\| \Mtilde - \Bl(M^*, \widehat{b}) \|_{B \cap E_2}^2 \notag \\ &=
\frac{|B|}{|B \cap E_2|} \| \Bl(M^* + W_2, \widehat{b}, E_2) -
\Bl(M^*, \widehat{b}) \|_{B \cap E_2}^2 \notag \\ &\le 2 \frac{|B|}{|B
  \cap E_2|} \Big( \| \Bl(M^* + W_2, \widehat{b}, E_2) - \Bl( \Bl(M^*,
\bhat) + W_2, \widehat{b}, E_2) \|_{B \cap E_2}^2 \notag\\ &\qquad
\qquad \qquad + \| \Bl( \Bl(M^*, \bhat) + W_2, \widehat{b}, E_2) -
\Bl(M^*, \widehat{b}) \|_{B \cap E_2}^2 \Big). \label{eq:case2-1}
\end{align}
Let us handle each term on the RHS of the last inequality
separately. First, by non-expansiveness of the projection operation
defined by equation \eqref{eq:projblock}, we have
\begin{align}
   \label{eq:proj}
\| \Bl(M^* + W_2, \widehat{b}, E_2) - \Bl( \Bl(M^*, \bhat) + W_2,
\widehat{b}, E_2) \|_{B \cap E_2}^2 \leq \| M^* - \Bl(M^*, \bhat)
\|^2_{B\cap E_2}.
\end{align}
We also require the following technical lemma:
\begin{lemma}
  \label{lem:blockcollision}
For any block $B$ and tuple $(i,j) \in B$, we have
\begin{align*}
\Pr \Big\{ (i,j) \in E_2 \;\; \big| \;\; |B \cap E_2| = k \Big\} =
\frac{k}{|B|}.
\end{align*}
\end{lemma}
\noindent See Section~\ref{sec:blockcollisionproof} for the proof of
this claim.

Returning to equation~\eqref{eq:proj} and taking expectation over the
randomness in $E_2$ (which, crucially, is independent of the
randomness in $\bhat$), we have
\begin{align}
& \EE_{E_2} \left[ \|M^* - \Bl(M^*, \bhat) \|_{B \cap E_2}^2 \mid \;
    |B \cap E_2| = k \right] \notag \\
& \qquad \qquad \qquad \qquad = \sum_{(i,j) \in B} \Pr \Big\{ (i,j)
  \in E_2 \;\; \big| \;\; |B \cap E_2| = k \Big\} \cdot \big[M^* -
    \Bl(M^*, \bhat) \big]_{ij}^2 \notag \\
& \qquad \qquad \qquad \qquad \stackrel{\2}{=} \sum_{(i,j) \in B}
  \frac{k}{|B|} \big[M^* - \Bl(M^*, \bhat) \big]_{ij}^2 \notag \\
& \qquad \qquad \qquad \qquad = \frac{k}{|B|} \|M^* - \Bl(M^*, \bhat)
  \|_B^2, \label{eq:case2-2}
\end{align}
where step $\2$ follows from Lemma~\ref{lem:blockcollision}. 

Additionally, notice that $[W_2]_{ij}$ for $(i, j) \in E_2$ is
independent and bounded within the interval $[-1, 1]$. Consequently,
we have
\begin{align}
\label{eq:case2-3}
\EE_{W_2} \left[ \| \Bl( \Bl(M^*, \bhat) + W_2, \widehat{b}, E_2) -
  \Bl(M^*, \widehat{b}) \|_{B \cap E_2}^2 \right] \leq 1,
\end{align}
where we have used the fact that the entries of the matrix $\Bl(M^*,
\bhat)$ are constant on the set of indices $B \cap E_2$.

It follows from equations~\eqref{eq:case2-1}, \eqref{eq:proj},
\eqref{eq:case2-2} and~\eqref{eq:case2-3} that
\begin{align*}
\EE \left[ \| \Mtilde - \Bl(M^*, \bhat) \|_B^2 \right] \leq 2 \EE
\left[ \frac{|B|}{|B \cap E_2|} \right] + 2 \EE \left[ \| M^* -
  \Bl(M^*, \bhat) \|_B^2 \right].
\end{align*}

Combining the two cases and summing over the blocks, we obtain that
\begin{align}
\EE \left[ \| \Mtilde - \Bl(M^*, \bhat) \|_F^2 \right] \leq 2 \sum_{B
  \in \mathcal{B}(\hat b)} \EE \left[ \frac{|B|}{|B \cap E_2| \vee 1}
  \right] + 2 \EE \left[ \| M^* - \Bl(M^*, \bhat) \|_F^2
  \right]. \label{eq:twoterms}
\end{align}
Note that the second term above is the same as the second term on the
RHS of inequality~\eqref{eq:sst-est}.

We now require the following definition, and two lemmas to complete
the proof. Given a matrix $M^*$ and a partition $C \in \chi_n$, define
its row average as
\begin{align*}
[\Rw(M^*, C)]_i = \frac{1}{|C(i)|} \sum_{j \in C(i)} M^*_j.
\end{align*}

\begin{lemma}
  \label{lem:invblock}
 With $S = \sum_{v \in V} 1/\sqrt{d_v}$ and for the partition $\bhat =
 \bl_{t} (\rspace(Y'_1))$, we have
\begin{align*}
\EE_{E_2} \left[ \sum_{B \in \mathcal{B}(\bhat)} \frac{|B|}{|B \cap
    E_2| \vee 1} \right] \leq n S.
\end{align*}
\end{lemma}

\begin{lemma}
  \label{lem:moncol}
Given any matrix $X \in [0,1]^{n \times n}$ with monotone columns, a
score vector $\rhat \in [0,n]^n$, and a value $t \in [0,n]$, we have
\begin{align*}
\| X - \Rw(X, \bl_t(\rhat) ) \|_F^2 \leq nt + 2 \| \rhat - \rspace(X)
\|_1.
\end{align*}
\end{lemma}

Applying Lemma~\ref{lem:invblock} with the expectation taken over the
edge set $E_2$ yields the desired bound on the first term of
inequality~\eqref{eq:twoterms}.

In order to bound the second term of inequality \eqref{eq:twoterms}, note that by definition, we have
\begin{align*}
\Bl(M^*, C) = \Rw( \Rw(M^*, C)^\top )^\top.
\end{align*}
Consequently, it holds that
\begin{align*}
\| M^* - \Bl(M^*, C) \|_F^2 &\leq 2\| M^* - \Rw(M^*, C) \|_F^2 + 2 \|
\Rw(M^*, C) - \Bl(M^*, C) \|_F^2 \\ &= 2\| M^* - \Rw(M^*, C) \|_F^2 +
2 \| \Rw(M^*, C)^\top - \Rw( \Rw(M^*, C)^\top, C) \|_F^2.
\end{align*}

Setting $C = \bl_S(\rhat)$ and applying Lemma \ref{lem:moncol} to both
the terms, we obtain
\begin{align*}
\| M^* - \Bl(M^*, \bl_S(\rhat) ) \|_F^2 \leq 2nS + 4 \| \rhat
- \rspace(M^*) \|_1.
\end{align*}
Applying Lemma~\ref{lem:scores} yields a bound on the second term in
expectation. This together with equations~\eqref{eq:sst-est}
and~\eqref{eq:twoterms} completes the proof of
Theorem~\ref{thm:sstupper} with the choice $t = \sum_{v \in V}
1/\sqrt{d_v}$.

\noindent It remains to prove Lemmas~\ref{lem:blockcollision},
\ref{lem:invblock} and~\ref{lem:moncol}.

%%%%%%%%%%%%%%%%%%%%%%%%%%%%%%%%%%%%%%%%%%%%%%%%%%%%%%%%%%%%%%%%%%%%%%%%%%%%%%%%

\subsubsection{Proof of Lemma~\ref{lem:blockcollision}}
\label{sec:blockcollisionproof}

Our proof relies crucially on the fact that one of the two sets is a
block.

For a fixed integer $k$, we condition on the event $\{|B \cap E_2| =
k\}$. Note that $E_2$ is the random edge set defined by
\begin{align*}
 E_2 = \pi(E) = \big\{ (i,j): (\pi(i),\pi(j)) \in E \big\},
\end{align*}
where $\pi$ is a uniform random permutation, and $E$ is a fixed
instance of $E_2$.  For any pair of tuples $(i, j), (k,\ell) \in B$,
consider the permutation $\tilde \pi$ defined by
\begin{itemize}
\item $\tilde \pi(i) = k$, $\tilde \pi(k) = i$, $\tilde \pi(j) = \ell$
  and $\tilde \pi(\ell) = j$;
\item $\tilde \pi(m) = m$ for $m \neq i,j,k$ or $\ell$.
\end{itemize}
Note that right-composition by $\tilde \pi$ is clearly a bijection
between the sets $\{\pi: (i,j) \in \pi(E) \}$ and $\{\pi: (k,\ell) \in
\pi(E) \}$.  Therefore, we have $|\{\pi: (i,j) \in E_2\}| = |\{\pi:
(k,\ell) \in E_2 \}|$. A counting argument then completes the
proof. Indeed, conditioned on the event $\{|B \cap E_2| = k\}$, we
have
\begin{align*}
\sum_{(i, j) \in B} \Pr \{(i,j) \in E_2 \} = \EE \Big[ \sum_{(i, j)
    \in B} \mathbf{1} \{(i,j) \in E_2 \} \Big] = k,
\end{align*}
which implies that $\Pr\{(i,j) \in E_2 \} = \frac{k}{|B|}$.

%%%%%%%%%%%%%%%%%%%%%%%%%%%%%%%%%%%%%%%%%%%%%%%%%%%%%%%%%%%%%%%%%
\subsubsection{Proof of Lemma~\ref{lem:invblock}}

Fix an individual block $B$ of dimensions $h \times w$, and let $E =
E_2$ for notational convenience. Define the random variable $Y = |B
\cap E| + 1$ so that $(|B \cap E| \vee 1)^{-1} \le 2/Y$.  Hence we
require a bound on the quantity $\EE[Y^{-1}]$. Toward this end, we
write
\begin{align*}
Y &= 1 + \sum_{(i,j) \in B} \mathbf{1}\{(i,j) \in E\} , \text{ and}
\\
Y^2 & = 1+ 2\sum_{(i,j) \in B} \mathbf{1}\{(i,j) \in E\} +
\sum_{(i,j),(i,j') \in B} \mathbf{1}\{(i,j),(i,j') \in E\} .
\end{align*}
Note that for $(i,j),(i',j') \in B$ where $i \ne i'$ and $j \ne j'$,
we have
\begin{align*}
\Pr\{(i,j) \in E\} & = \frac{2|E|}{n(n-1)}, \\ \Pr\{(i,j), (i,j') \in
E\} & = \frac{\sum_{v \in V} d_v(d_v - 1)}{n(n-1)(n-2)} , \text{ and}
\\
\Pr\{(i,j), (i',j') \in E\} & = \frac{4|E|^2 - 2 \sum_{v \in V}
  d_v(d_v - 1) - 2|E|}{n(n-1)(n-2)(n-3)}.
\end{align*}
Hence, we can compute the first two moments of $Y$ as
\begin{align*}
\EE[Y] &= 1 + \sum_{(i,j) \in B} \Pr\{(i,j) \in E\} = 1 + \frac{2 h w
  |E|}{n(n-1)} ,\text{ and} \\
\EE[Y^2] &= 1+ 2\sum_{(i,j) \in B} \Pr\{(i,j) \in E\} + \sum_{(i,j),
  (i',j') \in B} \Pr\{(i,j),(i',j') \in E\} \\ &= 1+ \frac{4 h w
  |E|}{n(n-1)} + \frac{2 h w |E|}{n(n-1)} + \Big[h w (w-1) + w h(h-1)
  \Big] \frac{\sum_{v \in V} d_v(d_v - 1)}{n(n-1)(n-2)} \\ &\quad \ +
h(h-1)w(w-1) \frac{4|E|^2 - 2 \sum_{v \in V} d_v(d_v - 1) -
  2|E|}{n(n-1)(n-2)(n-3)}.
\end{align*}
where for the last step we split into cases according to whether
$i=i'$ or $j=j'$. Therefore, the variance $\Var(Y)$ is equal to
\begin{align*} 
\EE[Y^2] - \EE[Y]^2 &= \frac{2 h w |E|}{n(n-1)} + \Big[h w (w-1) + w
  h(h-1) \Big] \frac{\sum_{v \in V} d_v(d_v - 1)}{n(n-1)(n-2)}
\\ &\quad \ + h(h-1)w(w-1) \frac{4|E|^2 - 2 \sum_{v \in V} d_v(d_v -
  1) - 2|E|}{n(n-1)(n-2)(n-3)} - \frac{4 h^2 w^2 |E|^2}{n^2(n-1)^2} .
\end{align*}
We note that
\begin{align*}
\frac{h(h-1)w(w-1)}{n(n-1)(n-2)(n-3)} - \frac{h^2 w^2}{n^2(n-1)^2} &=
\frac{hw[hw(4n - 6) - (h + w -1) n (n-1) ]}{n(n-1)(n-2)(n-3)} \\ &\leq
\frac{2 h^2 w^2}{n^2 (n-1)^2 (n-2) (n-3)}.
\end{align*}
where in the last step, we have used the fact that the quantity above
is maximized when $h = w$, and that $2 \leq h + w \leq n$ by the
construction of the blocks.

Combining the pieces, we conclude that $\Var(Y)$ is bounded by
\begin{align*} 
c \frac{h w |E|}{n^2} + c (h w^2 + w h^2) \frac{\sum_{v \in V}
  d_v^2}{n^3} + c \frac{h^2 w^2 |E|^2}{n^6} \le 2 c \frac{h w
  |E|}{n^2} + c (h w^2 + w h^2) \frac{\sum_{v \in V} d_v^2}{n^3}
\end{align*}
where the inequality holds because $h \le n$, $w \le n$ and $|E|\le
n^2$. Using the fact that $Y \ge 1$ and applying Chebyshev's
inequality, we obtain
\begin{align*}
\EE[Y^{-1}] &\le \Pr\Big\{Y \le \frac{\EE[Y]}2 \Big\} +
\frac{2}{\E[Y]} \\ & \le \frac{4}{\EE[Y]^2} \Var(Y) + \frac{2}{\E[Y]}
\\ & \le c \frac{n^4}{h^2 w^2 |E|^2} \Big[\frac{h w |E|}{n^2} + (h w^2
  + w h^2) \frac{\sum_{v \in V} d_v^2}{n^3} \Big] + c \frac{n^2}{h w
  |E|} \\ & = 2 c \frac{n^2}{h w |E|} + c n \frac{h+w}{h w}
\frac{\sum_{v \in V} d_v^2}{|E|^2}.
\end{align*}

Now the above bound yields
\begin{align*}
\EE \frac{|B|}{Y} \le 2 c \frac{n^2}{|E|} + c n (h+w) \frac{\sum_{v
    \in V} d_v^2}{|E|^2}.
\end{align*}
Note that there are at most $m^2 = (n / S)^2$ blocks in total and the
sum of $h$ over $m-1$ off-diagonal blocks vertically is bounded by $n$
(similarly for $w$). Thus we conclude that
\begin{align*}
  \EE \sum_{B \in \mathcal{B}(\bhat)} \frac{|B|}{|B \cap E| \vee 1}
  \le c \frac{m^2 n^2}{|E|} + c \, m n^2 \frac{\sum_{v \in V}
    d_v^2}{|E|^2}.
\end{align*}

In order to complete the proof, it suffices to show that
\begin{align*}
  \frac{n^2}{|E|} \Big(\sum_{v \in V} \frac{1}{\sqrt{d_v}} \Big)^{-2}
  + n \Big(\sum_{v \in V} \frac{1}{\sqrt{d_v}} \Big)^{-1}
  \frac{\sum_{v \in V} d_v^2}{|E|^2} \le \frac cn \sum_{v \in V}
  \frac{1}{\sqrt{d_v}}.
\end{align*}
Note that Lemma~\ref{lem:rearrange} implies that
\begin{align*}
  2|E| \Big( \sum_{v \in V} \frac 1{\sqrt{d_v}} \Big)^2 = \Big(
  \sum_{v \in V} d_v \Big) \Big( \sum_{v \in V} \frac 1{\sqrt{d_v}}
  \Big)^2 \ge n^3.
\end{align*}
It follows that
\begin{align*}
  \frac{n^2}{|E|} \Big(\sum_{v \in V} \frac{1}{\sqrt{d_v}} \Big)^{-2}
  \le \frac 2n \le \frac 2n \sum_{v \in V}
  \frac{1}{\sqrt{d_v}},
\end{align*}
and that
\begin{align*}
  n \Big(\sum_{v \in V} \frac{1}{\sqrt{d_v}} \Big)^{-1} \frac{\sum_{v
      \in V} d_v^2}{|E|^2} \le \frac{4}{n^2} \frac{\sum_{v \in V}
    d_v^2}{\sum_{v\in V} d_v} \Big(\sum_{v \in V} \frac{1}{\sqrt{d_v}}
  \Big) \le \frac 4n \sum_{v \in V} \frac{1}{\sqrt{d_v}}
\end{align*}
since $d_v \le n$.

%%%%%%%%%%%%%%%%%%%%%%%%%%%%%%%%%%%%%%%%%%%%%%%%%%%%%%%%%%%%%%%%%%%%%%%%%%%%%%%%

\subsubsection{Proof of Lemma~\ref{lem:moncol}}
This lemma is a generalization of an approximation theorem due to
Chatterjee~\cite{chatterjee15matrix} and Shah et
al.~\cite{shah17stochastically} to the noisy and two-dimensional
setting.

We use the shorthand $\Chatt = \bl_t(\rhat)$ for the rest of the
proof. Also define the set of placeholder elements in the partition
$\Chatt$ as
\begin{align*}
\elts(\Chatt) = \{i : i \text{ is smallest index in some set }I \in \Chatt \}.
\end{align*}
We are now ready to prove the lemma. Begin by writing
\begin{align*}
\| X - \Rw( X, \Chatt) \|_F^2 &= \sum_{k=1}^n \Big\| X_k -
\frac{1}{|\Chatt(k)|} \sum_{j \in \Chatt(k)} X_j \Big\|_2^2
\\
& \stackrel{\1}{\leq} \sum_{k=1}^n \Big\| X_k - \frac{1}{|\Chatt(k)|}
\sum_{j \in \Chatt(k)} X_j \Big\|_1 \\
& \stackrel{\2}{\leq} \sum_{k=1}^n \frac{1}{|\Chatt(k)|} \sum_{j \in
  \Chatt(k)} \| X_k - X_j \|_1 \\ &\stackrel{\3}{\leq} \sum_{k=1}^n
\frac{1}{|\Chatt(k)|} \sum_{j \in \Chatt(k)} | \rspace(X)_k
- \rspace(X)_j | \\
& = \sum_{k \in \elts(\Chatt)} \frac{1}{|\Chatt(k)|} \sum_{i \in
  \Chatt(k)} \sum_{j \in \Chatt(k)} | \rspace(X)_i - \rspace(X)_j | \\
& \leq \sum_{k \in \elts(\Chatt)} \frac{1}{|\Chatt(k)|} \sum_{i,j \in
  \Chatt(k)} \big( | \rhat_i - \rspace(X)_i | + | \rhat_j
- \rspace(X)_j | + | \rhat_i - \rhat_j | \big) \\ &\stackrel{\4}{\leq}
\| \rhat - \rspace(X) \|_1 + \| \rhat - \rspace(X) \|_1 + \sum_{k \in
  \elts(\Chatt)} t |\Chatt(k) | \\
& = 2 \| \rhat - \rspace(X) \|_1 + n t.
\end{align*}
Step $\1$ follows from the fact that each entry of the difference
matrix $X - \Rw( X, \Chatt)$ is bounded in the interval $[-1,1]$; step
$\2$ follows from Jensen's inequality and convexity of the $\ell_1$
norm; step $\3$ uses the fact that for fixed $k$ and $j$, the quantity
$X_{k\ell} - X_{j\ell}$ has the same sign for all $\ell \in [n]$ due
to the monotonicity of columns of the matrix $X$; step $\4$ uses the
property of the blocking partition $\Chatt$, which ensures that
$|\rhat_i - \rhat_j| \leq t$ when the inclusion $i, j \in \Chatt(k)$
is satisfied for some $k$.  This completes the proof.

% Local Variables:
% TeX-master: "worstavg_design_arxiv"
% End: 

\section{Discussion}

In this paper, we studied the problem of estimating the comparison
probabilities from noisy pairwise comparisons under worst-case and
average-case design assumptions. We exhibited a dichotomy between
worst-case and average-case models for permutation-based models, which
suggests that a similar distinction may exist even for their
parametric counterparts. Our bounds leave a few interesting questions
unresolved: Is there a sharp characterization of the diameter
$\mathcal{A}(G)$ quantifying the approximation error of a comparison
topology $G$? The Borda count estimator, a variant of which we
analyzed, is known to achieve a sub-optimal rate in the case of full
observations; the estimator of Braverman and
Mossel~\cite{braverman08noisy} achieves the optimal rate over the
noisy sorting class. What is the analog of such an estimator in the
average-case setting with partial pairwise comparisons? Is there a
computational lower bound to show that our estimators are the best
possible polynomial-time algorithms for SST matrix estimation in the
average-case setting?

\subsection*{Acknowledgements}

This work was partially supported by National Science Foundation grants
NSF-DMS-1612948, CCF-1528132, and CCF-0939370 (Science of Information), and DOD Advanced Research Projects Agency grant
W911NF-16-1-0552.
CM was supported in part by NSF CAREER DMS-1541099 and was visiting the Simons Institute for the Theory of Computing while this work was done.

\appendix
\section{Bounds on the minimax denoising error} \label{app:denoise}

As we saw in Theorem \ref{thm:minimax}, the minimax risk of Frobenius
norm estimation is prohibitively large for many comparison
topologies. In some applications, however, it may be of interest to
control the denoising error, which is the error we make on the
observations seen on the edges of the graph. Accordingly, we define
the quantity
\begin{align*}
\Erisk(G, \Cmodel) = \inf_{\Mhat = f(Y(G))} \sup_{M^* \in \Cmodel} \EE \Big[ \frac{1}{|E|}\|\Mhat - M^* \|_E^2 \Big],
\end{align*}
where we have used a normalization of $|E|$ to provide an average
entry-wise bound on the denoising error.  The following theorem
provides bounds on the minimax denoising error for fixed topologies.

\begin{theorem}
  \label{thm:est}
For any connected graph $G$, we have
\begin{align}
\Erisk(G, \Cns) \geq \frac{c_1}{|E|} \max_{S \in \mathcal{C}_G}
\frac{|V(S)|^2}{|E(S)|}, \quad \text{ and} \quad \Erisk(G, \csst) &
\leq \frac{c_2 n \log^2 n}{|E|} .
\end{align}
\end{theorem}

Again, the lower bound on the error of the noisy sorting class
provides a lower bound for the SST class. Conversely, the upper bound
on the error for the SST class upper bounds the error for the noisy
sorting class.

For many graphs used in practice, the lower bound can be evaluated to
show that Theorem \ref{thm:est} provides a sharp characterization of
the denoising error up to logarithmic factors.

The upper bound is obtained by the least squares estimator
\begin{align*}
\Mhat_{{\sf LS}} = \arg \min_{\Mhat \in \csst} \|Y - M^*\|_E^2.
\end{align*}
While we do not know yet whether such an estimator is computable in
polynomial time, analyzing it provides a notion of the fundamental
limits of the problem. In particular, it is clear that the denoising
problem is easier than Frobenius norm estimation, and we obtain
consistent rates provided that the number of edges in the graph
satisfies $|E| = \omega(n \log^2 n)$.

%%%%%%%%%%%%%%%%%%%%%%%%%%%%%%%%%%%%%%%%%%%%%%%%%%%%%%%%%%%%%%%%%%%%%%%%%%%%%%
\subsection{Proof of Theorem~\ref{thm:est}}

In this section, we prove Theorem~\ref{thm:est} on the denoising error
rate of the problem, splitting it into proofs of the lower and upper
bounds.

%%%%%%%%%%%%%%%%%%%%%%%%%%%%%%%%%%%%%%%%%%%%%%%%%%%%%%%%%%%%%%%%%%%%%%%%

\subsubsection{Proof of lower bound}

In order to prove the lower bound, we construct a suitable local
packing $\mathcal{P}$ of the parameter space $\Cns$, and then apply
Fano's inequality. For simpler presentation, we describe the packing
$\mathcal{P}$ by gradually putting constraints on its members. First,
every matrix in $\mathcal{P}$ is chosen to be $\Mns(\pi, \lambda)$ for
a fixed $\lambda$ and some permutation $\pi$, so we focus on selecting
the permutations $\pi$.

Consider any connected subgraph $S \in \mathcal{C}_G$ with at least
two vertices.  Let the vertices of $S$ form the top $|V(S)|$ items and
choose the same ranking for the vertices of $S^c$ for each instance in
the packing. Then all the matrices in the packing $\mathcal{P}$ have
the same $(i,j)$-th entry if $i \in S^c$ or $j \in S^c$. Hence the KL
divergence between any two models with underlying matrices in the
packing $\mathcal{P}$ is bounded by $9 \lambda^2 |E(S)|$, by Lemma
\ref{lem:nskl}.

Next, fix a spanning tree $T(S)$ of $S$ which has $|V(S)|-1$
edges. Note that all the $2^{|V(S)|-1}$ assignments of values to these
edges
\begin{align*}
\{M_{ij}: (i,j) \in T(S), \, i<j \} \in \{1/2 + \lambda, 1/2 - \lambda \}^{|V(S)|-1}
\end{align*} 
are possible, since there are no cycle conflicts in the spanning
tree. Using the Gilbert-Varshamov bound, we are guaranteed that there
are constants $a$ and $b$ such that at least $2^{a |V(S)|}$ such
assignments are separated pairwise by $b |V(S)|$ in the Hamming
distance. We choose the packing $\mathcal{P}$ consisting of matrices
corresponding to these assignments, so that $\|M-M'\|_F^2 \ge 8 b
\lambda^2 |V(S)|$ for any distinct $M,M' \in \mathcal{P}$.

Finally, Fano's inequality implies that
\begin{align*}
|E|\, \Erisk(G, \Cns) \geq 8 b \lambda^2 |V(S)| \left(1 - \frac{9
  \lambda^2 |E(S)| + \log 2}{a |V(S)|} \right).
\end{align*}
The proof then follows by choosing $\lambda^2 = c
\frac{|V(S)|}{|E(S)|}$, for a sufficiently small constant $c$.  \qed

%%%%%%%%%%%%%%%%%%%%%%%%%%%%%%%%%%%%%%%%%%%%%%%%%%%%%%%%%%%%%%%%%%%%%%%%%%%

\subsubsection{Proof of upper bound}

As mentioned before, we obtain the upper bound by considering the
estimator $\Mhat_{{\sf LS}}$. The proof follows from previous results
on the full observation case~\cite{shah17stochastically}, but we
provide it for completeness. Note that for each $(i, j) \in E$, the
observation model takes the form
\begin{align*}
Y_{ij} = M^*_{ij} + W_{ij},
\end{align*}
where $W_{ij}$ is a zero-mean noise variable lying in the interval
$[-1,1]$.

The optimality of $\Mhat_{{\sf LS}}$ and feasibility of $M^*$ imply
that we must have the basic inequality \mbox{$\|Y - \Mhat_{{\sf
      LS}}\|_E^2 \le \|Y - M^*\|_E^2$,} which after simplification,
leads to
\begin{align}
  \label{EqnEarlyBasic}
\frac{1}{2} \|\Delta\|^2_{E} \leq \langle \Delta, W \rangle_E,
\end{align}
where $\Delta = \Mhat_{\sf LS} - M^*$, and $\langle A, B \rangle_E =
\sum_{(i,j) \in E} A_{ij} B_{ij}$ denotes the trace inner product
restricted to the indices in $E$.

In order to establish the upper bound, we first define the class of
difference matrices $\cdiff \defn \{ M - M' \mid M, M' \in \csst\}$,
as well as the associated random variable
\begin{align*}
Z(t) \defn \sup_{D \in \cdiff: \|D \|_E \leq t} \langle D, W
\rangle_E.
\end{align*}
With this notation, inequality~\eqref{EqnEarlyBasic} implies
$\frac{1}{2} \|\Delta\|^2_{E} \leq Z(\|\Delta\|_E)$.  It follows from
the star-shaped property\footnote{A set $S$ is said to be star-shaped
  if $t \in S$ implies that $\alpha t \in S \text{ for all } \alpha
  \in [0,1]$} of the set $\cdiff$ that the following critical
inequality is satisfied for some $\delta > 0$:
\begin{align*}
\EE [Z(\delta)] \leq \frac{\delta^2}{2}.
\end{align*}
We are interested in the smallest such value $\delta$. In order to
find it, we use Dudley's entropy integral, for which we require a
bound on the covering number of the class $\cdiff$. Such a bound was
calculated for the Frobenius norm by Shah et
al.~\cite{shah17stochastically} using the results of Gao and
Wellner~\cite{gao2007entropy}. Clearly, since $\|M_i - M_j \|_E^2 \leq
\|M_i - M_j \|_F^2$, a $\delta$-covering in the Frobenius norm
automatically serves as a $\delta$-covering in the edge norm $\| \cdot
\|_E$. Thus, we have the following lemma.

%%%%%%%%%%%%%%%%%%%%%%%%%%%%%%%%%%%%%%%%%%%%%%%%%%%%%%%%%%%%%%%%%%%%%%%%%%%%%%
\begin{lemma}\cite{shah17stochastically}
  \label{lem:covering}
For every $\epsilon > 0$, we have the metric entropy bound
\begin{align*}
\log N(\epsilon, \cdiff, \| \cdot \|_E ) \leq \log N(\epsilon, \cdiff,
\| \cdot \|_F ) \leq 9 \frac{n^2}{\epsilon^2} \left(\log
\frac{n}{\epsilon}\right)^2 + 9 n \log n.
\end{align*}
\end{lemma}

Dudley's entropy integral then yields that for all $t >0$, we have
\begin{align*}
\EE[Z(t)] & \leq c \inf_{\delta \in [0,n]} \Big\{ n\delta +
\int_{\delta/2}^t \sqrt{\log N(\epsilon, \cdiff \cap \mathbb{B}_E(t),
  \|\cdot \|_E)} d \epsilon \Big\} \\ &\leq c \Big\{ n^{-8} +
\int_{n^{-9}/2}^t \sqrt{\log N(\epsilon, \cdiff, \| \cdot \|_E)}
d\epsilon \Big\}.
\end{align*}
After some algebra (for details, see Shah et
al.~\cite{shah17stochastically}), we have
\begin{align*}
\EE[Z(t)] &\leq c \big\{ n \log^2 n + t \sqrt{n \log n} \big\}.
\end{align*}
Setting $t = c \sqrt{n} \log n$ completes the proof. \qed

\bibliographystyle{alpha}
\bibliography{fixed_comparison}

\newcommand{\etalchar}[1]{$^{#1}$}
\begin{thebibliography}{CBCTH13}

\bibitem[BA99]{barabasi1999emergence}
Albert-L{\'a}szl{\'o} Barab{\'a}si and R{\'e}ka Albert.
\newblock Emergence of scaling in random networks.
\newblock {\em Science}, 286(5439):509--512, 1999.

\bibitem[Bar03]{barnett2003modern}
William Barnett.
\newblock The modern theory of consumer behavior: Ordinal or cardinal?
\newblock {\em Quarterly Journal of Austrian Economics}, 6(1):41--65, 2003.

\bibitem[BDPR84]{bril1984algorithm}
Gordon Bril, Richard Dykstra, Carolyn Pillers, and Tim Robertson.
\newblock Algorithm {AS} 206: isotonic regression in two independent variables.
\newblock {\em Journal of the Royal Statistical Society. Series C (Applied
  Statistics)}, 33(3):352--357, 1984.

\bibitem[BM08]{braverman08noisy}
Mark Braverman and Elchanan Mossel.
\newblock Noisy sorting without resampling.
\newblock In {\em Proceedings of the nineteenth annual ACM-SIAM symposium on
  Discrete algorithms}, pages 268--276. Society for Industrial and Applied
  Mathematics, 2008.

\bibitem[BM10]{barg2010codes}
Alexander Barg and Arya Mazumdar.
\newblock Codes in permutations and error correction for rank modulation.
\newblock {\em IEEE Transactions on Information Theory}, 56(7):3158--3165,
  2010.

\bibitem[BM15]{bardenet15concentration}
R\'emi Bardenet and Odalric-Ambrym Maillard.
\newblock Concentration inequalities for sampling without replacement.
\newblock {\em Bernoulli}, 21(3):1361--1385, 2015.

\bibitem[BMR10]{baltrunas2010group}
Linas Baltrunas, Tadas Makcinskas, and Francesco Ricci.
\newblock Group recommendations with rank aggregation and collaborative
  filtering.
\newblock In {\em Proceedings of the Fourth ACM Conference on Recommender
  systems}, pages 119--126. ACM, 2010.

\bibitem[BT52]{bradley52rank}
Ralph~A. Bradley and Milton~E. Terry.
\newblock Rank analysis of incomplete block designs. {I}. {T}he method of
  paired comparisons.
\newblock {\em Biometrika}, 39:324--345, 1952.

\bibitem[BW97]{ballinger1997decisions}
T.~Parker Ballinger and Nathaniel~T. Wilcox.
\newblock Decisions, error and heterogeneity.
\newblock {\em The Economic Journal}, 107(443):1090--1105, 1997.

\bibitem[Cat12]{cattelan2012models}
Manuela Cattelan.
\newblock Models for paired comparison data: A review with emphasis on
  dependent data.
\newblock {\em Statistical Science}, pages 412--433, 2012.

\bibitem[CBCTH13]{chen2013pairwise}
Xi~Chen, Paul~N. Bennett, Kevyn Collins-Thompson, and Eric Horvitz.
\newblock Pairwise ranking aggregation in a crowdsourced setting.
\newblock In {\em Proceedings of the sixth ACM international conference on Web
  search and data mining}, pages 193--202. ACM, 2013.

\bibitem[CGMS17]{chen2017competitive}
Xi~Chen, Sivakanth Gopi, Jieming Mao, and Jon Schneider.
\newblock Competitive analysis of the top-k ranking problem.
\newblock In {\em Proceedings of the Twenty-Eighth Annual ACM-SIAM Symposium on
  Discrete Algorithms}, pages 1245--1264. SIAM, 2017.

\bibitem[Cha15]{chatterjee15matrix}
Sourav Chatterjee.
\newblock Matrix estimation by universal singular value thresholding.
\newblock {\em Ann. Statist.}, 43(1):177--214, 2015.

\bibitem[CM16]{chatterjee16estimation}
Sabyasachi Chatterjee and Sumit Mukherjee.
\newblock On estimation in tournaments and graphs under monotonicity
  constraints.
\newblock {\em arXiv preprint arXiv:1603.04556}, 2016.

\bibitem[CN91]{caplin1991aggregation}
Andrew Caplin and Barry Nalebuff.
\newblock Aggregation and social choice: a mean voter theorem.
\newblock {\em Econometrica: Journal of the Econometric Society}, pages 1--23,
  1991.

\bibitem[CS15]{chen2015spectral}
Yuxin Chen and Changho Suh.
\newblock Spectral mle: Top-k rank aggregation from pairwise comparisons.
\newblock In {\em International Conference on Machine Learning}, pages
  371--380, 2015.

\bibitem[DG77]{diaconis1977spearman}
Persi Diaconis and Ronald~L. Graham.
\newblock Spearman's footrule as a measure of disarray.
\newblock {\em Journal of the Royal Statistical Society. Series B
  (Methodological)}, pages 262--268, 1977.

\bibitem[DKNS01]{dwork2001rank}
Cynthia Dwork, Ravi Kumar, Moni Naor, and Dandapani Sivakumar.
\newblock Rank aggregation methods for the web.
\newblock In {\em Proceedings of the 10th International Conference on World
  Wide Web}, pages 613--622. ACM, 2001.

\bibitem[DM56]{de1756doctrine}
Abraham De~Moivre.
\newblock {\em The doctrine of chances: or, A method of calculating the
  probabilities of events in play}, volume~1.
\newblock Chelsea Publishing Company, 1756.

\bibitem[Fis73]{fishburn1973binary}
Peter~C. Fishburn.
\newblock Binary choice probabilities: on the varieties of stochastic
  transitivity.
\newblock {\em Journal of Mathematical psychology}, 10(4):327--352, 1973.

\bibitem[FMR16]{flammarion16optimal}
Nicolas Flammarion, Cheng Mao, and Philippe Rigollet.
\newblock Optimal rates of statistical seriation.
\newblock {\em arXiv preprint arXiv:1607.02435}, 2016.

\bibitem[FV93]{fligner1993probability}
Michael~A. Fligner and Joseph~S. Verducci.
\newblock {\em Probability models and statistical analyses for ranking data},
  volume~80.
\newblock Springer, 1993.

\bibitem[GW07]{gao2007entropy}
Fuchang Gao and Jon~A. Wellner.
\newblock Entropy estimate for high-dimensional monotonic functions.
\newblock {\em Journal of Multivariate Analysis}, 98(9):1751--1764, 2007.

\bibitem[Hak62]{hakimi1962realizability}
S.~Louis Hakimi.
\newblock On realizability of a set of integers as degrees of the vertices of a
  linear graph. i.
\newblock {\em Journal of the Society for Industrial and Applied Mathematics},
  10(3):496--506, 1962.

\bibitem[Hav55]{havel1955remark}
V{\'a}clav Havel.
\newblock A remark on the existence of finite graphs.
\newblock {\em Casopis Pest. Mat.}, 80:477--480, 1955.

\bibitem[HMG06]{herbrich2006trueskill}
Ralf Herbrich, Tom Minka, and Thore Graepel.
\newblock Trueskill™: a {B}ayesian skill rating system.
\newblock In {\em Proceedings of the 19th International Conference on Neural
  Information Processing Systems}, pages 569--576. MIT Press, 2006.

\bibitem[Hoe63]{hoeffding1963probability}
Wassily Hoeffding.
\newblock Probability inequalities for sums of bounded random variables.
\newblock {\em Journal of the American statistical association},
  58(301):13--30, 1963.

\bibitem[HOX14]{hajek2014minimax}
Bruce Hajek, Sewoong Oh, and Jiaming Xu.
\newblock Minimax-optimal inference from partial rankings.
\newblock In {\em Advances in Neural Information Processing Systems}, pages
  1475--1483, 2014.

\bibitem[HSRW16]{heckel2016active}
Reinhard Heckel, Nihar~B. Shah, Kannan Ramchandran, and Martin~J. Wainwright.
\newblock Active ranking from pairwise comparisons and when parametric
  assumptions don't help.
\newblock {\em arXiv preprint arXiv:1606.08842}, 2016.

\bibitem[JKSO16]{jang2016top}
M.~Jang, S.~Kim, C.~Suh, and S.~Oh.
\newblock Top-$ k $ ranking from pairwise comparisons: When spectral ranking is
  optimal.
\newblock {\em arXiv preprint arXiv:1603.04153}, 2016.

\bibitem[JN11]{jamieson2011active}
Kevin~G. Jamieson and Robert~D. Nowak.
\newblock Active ranking using pairwise comparisons.
\newblock In {\em Advances in Neural Information Processing Systems}, pages
  2240--2248, 2011.

\bibitem[Ken48]{kendall1948rank}
Maurice~G. Kendall.
\newblock {\em Rank correlation methods.}
\newblock Charles Griffin and Company, London, 1948.

\bibitem[KO16]{khetan2016data}
Ashish Khetan and Sewoong Oh.
\newblock Data-driven rank breaking for efficient rank aggregation.
\newblock {\em Journal of Machine Learning Research}, 17(193):1--54, 2016.

\bibitem[KTT15]{kiraly2015algebraic}
Franz~J. Kir{\'a}ly, Louis Theran, and Ryota Tomioka.
\newblock The algebraic combinatorial approach for low-rank matrix completion.
\newblock {\em Journal of Machine Learning Research}, 16:1391--1436, 2015.

\bibitem[Luc59]{luce59individual}
R.~Duncan Luce.
\newblock {\em Individual choice behavior: {A} theoretical analysis}.
\newblock John Wiley \& Sons, Inc., New York; Chapman \& Hall, Ltd., London,
  1959.

\bibitem[Mah00]{mahmoud00sorting}
H.~M. Mahmoud.
\newblock {\em Sorting: A Distribution Theory}.
\newblock Wiley Series in Discrete Mathematics and Optimization. Wiley, 2000.

\bibitem[Mar96]{marden1996analyzing}
John~I. Marden.
\newblock {\em Analyzing and modeling rank data}.
\newblock CRC Press, 1996.

\bibitem[MG15]{maystre2015robust}
Lucas Maystre and Matthias Grossglauser.
\newblock Robust active ranking from sparse noisy comparisons.
\newblock {\em arXiv preprint arXiv:1502.05556}, 2015.

\bibitem[ML65]{mclaughlin1965stochastic}
Don~H. McLaughlin and R.~Duncan Luce.
\newblock Stochastic transitivity and cancellation of preferences between
  bitter-sweet solutions.
\newblock {\em Psychonomic Science}, 2(1-12):89--90, 1965.

\bibitem[NOS16]{negahban2016rank}
Sahand Negahban, Sewoong Oh, and Devavrat Shah.
\newblock Rank centrality: Ranking from pairwise comparisons.
\newblock {\em Operations Research}, 2016.

\bibitem[NOTX17]{negahban2017learning}
Sahand Negahban, Sewoong Oh, Kiran~K Thekumparampil, and Jiaming Xu.
\newblock Learning from comparisons and choices.
\newblock {\em arXiv preprint arXiv:1704.07228}, 2017.

\bibitem[NP66]{neyman1966joint}
Jerzy Neyman and Egon~S. Pearson.
\newblock {\em Joint statistical papers}.
\newblock Univ of California Press, 1966.

\bibitem[PABN16]{pimentel2016characterization}
Daniel~L. Pimentel-Alarc{\'o}n, Nigel Boston, and Robert~D. Nowak.
\newblock A characterization of deterministic sampling patterns for low-rank
  matrix completion.
\newblock {\em IEEE Journal of Selected Topics in Signal Processing},
  10(4):623--636, 2016.

\bibitem[PNZ{\etalchar{+}}15]{park2015preference}
Dohyung Park, Joe Neeman, Jin Zhang, Sujay Sanghavi, and Inderjit Dhillon.
\newblock Preference completion: Large-scale collaborative ranking from
  pairwise comparisons.
\newblock In {\em International Conference on Machine Learning}, pages
  1907--1916, 2015.

\bibitem[PWC16]{pananjady2016linear}
Ashwin Pananjady, Martin~J. Wainwright, and Thomas~A. Courtade.
\newblock Linear regression with an unknown permutation: Statistical and
  computational limits.
\newblock {\em arXiv preprint arXiv:1608.02902}, 2016.

\bibitem[PWC17]{pananjady2017denoising}
Ashwin Pananjady, Martin~J. Wainwright, and Thomas~A. Courtade.
\newblock Denoising linear models with permuted data.
\newblock {\em arXiv preprint arXiv:1704.07461}, 2017.

\bibitem[RA16]{rajkumar2016when}
Arun Rajkumar and Shivani Agarwal.
\newblock When can we rank well from comparisons of $o(n\log(n))$ non-actively
  chosen pairs?
\newblock In {\em 29th COLT}, volume~49, pages 1376--1401, 2016.

\bibitem[SBB{\etalchar{+}}16]{shah16estimation}
Nihar~B. Shah, Sivaraman Balakrishnan, Joseph Bradley, Abhay Parekh, Kannan
  Ramchandran, and Martin~J. Wainwright.
\newblock Estimation from pairwise comparisons: sharp minimax bounds with
  topology dependence.
\newblock {\em J. Mach. Learn. Res.}, 17:Paper No. 58, 47, 2016.

\bibitem[SBC05]{stewart2005absolute}
Neil Stewart, Gordon D.~A. Brown, and Nick Chater.
\newblock Absolute identification by relative judgment.
\newblock {\em Psychological Review}, 112(4):881, 2005.

\bibitem[SBGW17]{shah17stochastically}
Nihar~B. Shah, Sivaraman Balakrishnan, Adityanand Guntuboyina, and Martin~J.
  Wainwright.
\newblock Stochastically transitive models for pairwise comparisons:
  statistical and computational issues.
\newblock {\em IEEE Trans. Inform. Theory}, 63(2):934--959, 2017.

\bibitem[SBW16a]{shah16feeling}
Nihar~B. Shah, Sivaraman Balakrishnan, and Martin~J. Wainwright.
\newblock Feeling the {B}ern: Adaptive estimators for {B}ernoulli probabilities
  of pairwise comparisons.
\newblock In {\em Information Theory (ISIT), 2016 IEEE International Symposium
  on}, pages 1153--1157. IEEE, 2016.

\bibitem[SBW16b]{shah2016permutation}
Nihar~B. Shah, Sivaraman Balakrishnan, and Martin~J. Wainwright.
\newblock A permutation-based model for crowd labeling: Optimal estimation and
  robustness.
\newblock {\em arXiv preprint arXiv:1606.09632}, 2016.

\bibitem[SW15]{shah2015simple}
Nihar~B. Shah and Martin~J. Wainwright.
\newblock Simple, robust and optimal ranking from pairwise comparisons.
\newblock {\em arXiv preprint arXiv:1512.08949}, 2015.

\bibitem[Thu27]{thurstone27law}
Louis~L. Thurstone.
\newblock A law of comparative judgment.
\newblock {\em Psychological review}, 34(4):273, 1927.

\bibitem[Vin90]{vince90a}
A.~Vince.
\newblock A rearrangement inequality and the permutahedron.
\newblock {\em Amer. Math. Monthly}, 97(4):319--323, 1990.

\bibitem[WJJ13]{wauthier2013efficient}
Fabian Wauthier, Michael Jordan, and Nebojsa Jojic.
\newblock Efficient ranking from pairwise comparisons.
\newblock In {\em International Conference on Machine Learning}, pages
  109--117, 2013.

\bibitem[Yi04]{yi04theta}
Jinhee Yi.
\newblock Theta-function identities and the explicit formulas for
  {T}heta-function and their applications.
\newblock {\em J. Math. Anal. Appl.}, 292(2):381--400, 2004.

\end{thebibliography}

\end{document}